\documentclass{article}

\usepackage[final]{neurips_2023}

\usepackage[utf8]{inputenc} 
\usepackage[T1]{fontenc}    
\usepackage{hyperref}       
\usepackage{url}            
\usepackage{booktabs}       
\usepackage{amsfonts, amsmath}       
\usepackage{nicefrac}       
\usepackage{microtype}      
\usepackage{xcolor}         
\usepackage{amsmath}
\usepackage{enumitem}
\usepackage{graphicx}
\usepackage{multirow}
\usepackage{xcolor}

\usepackage{caption}

\allowdisplaybreaks

\usepackage{booktabs}
\newcommand{\ra}[1]{\renewcommand{\arraystretch}{#1}}

\usepackage{bm}

\hypersetup{
    colorlinks = true,
    citecolor = {rgb, 255:red, 160; green, 0; blue, 0 }
}

\title{Structure Learning with Adaptive Random Neighborhood Informed MCMC}

\author{%
  Alberto Caron\thanks{These authors contributed equally to this work.} 
    \\
  The Alan Turing Institute\\
  London, UK\\
  \texttt{acaron@turing.ac.uk} \\
  \And
  Xitong Liang\footnotemark[1] \\
  Department of Statistical Sciences\\
  University College London \\
  London, UK\\
  \texttt{xitong.liang.18@ucl.ac.uk} \\
  \AND
  Samuel Livingstone \\
  Department of Statistical Sciences\\
  University College London \\
  London, UK\\
  \texttt{samuel.livingstone@ucl.ac.uk} \\
  \And
  Jim Griffin \\
  Department of Statistical Sciences\\
  University College London \\
  London, UK \\
  \texttt{j.griffin@ucl.ac.uk} \\
}

\begin{document}

\maketitle
\setcounter{footnote}{0} 

\begin{abstract}
     In this paper, we introduce a novel MCMC sampler, PARNI-DAG, for a fully-Bayesian approach to the problem of structure learning under observational data. Under the assumption of causal sufficiency, the algorithm allows for approximate sampling directly from the posterior distribution on Directed Acyclic Graphs (DAGs). PARNI-DAG performs efficient sampling of DAGs via \textit{locally informed}, \textit{adaptive random neighborhood} proposal that results in better mixing properties. In addition, to ensure better scalability with the number of nodes, we couple PARNI-DAG with a pre-tuning procedure of the sampler's parameters that exploits a skeleton graph derived through some constraint-based or scoring-based algorithms. Thanks to these novel features, PARNI-DAG quickly converges to high-probability regions and is less likely to get stuck in local modes in the presence of high correlation between nodes in high-dimensional settings. After introducing the technical novelties in PARNI-DAG, we empirically demonstrate its mixing efficiency and accuracy in learning DAG structures on a variety of experiments.\footnote{Code to implement the PARNI-DAG proposal and replicate the experimental sections is available at \url{https://github.com/XitongLiang/The-PARNI-scheme/tree/main/Structure-Learning}.}
    
\end{abstract}

\section{Introduction} \label{sec1:intro}

Structure Learning and Causal Discovery are concerned with reconstructing a graphical model underlying a set of random variables from data, in the form of a Directed Acyclic Graph (DAG), provided that causal identifiability assumptions hold \citep{pearl_2009_book, drton_2017, glymour_2019}. Causal relations can typically be inferred if randomized experiments are carried out \citep{pearl_2009_book}. However, in many relevant applied fields (\textit{e.g.} biology, genomics, ecology, etc.), only observational data are accessible, as performing interventions is usually costly or simply unfeasible. Structure learning from observational data is a challenging problem, with the space of possible DAGs growing super-exponentially in the number of nodes \citep{robinson_1977}.

\paragraph{Related work.} Early contributions on structure learning algorithms include \textit{constraint-based methods} \citep{spirtes_2000_book, pearl_2009_book}, such as PC and FCI, that utilize conditional independence testing to output a Markov Equivalence Class (MEC). Another stream of literature focuses on \textit{score-based methods} \citep{geiger_1994}, such as Greedy Equivalent Search (GES) \citep{chickering_1996, chickering_2002}, which define and consequently maximize a score function associated with different MECs. Alternatively, \textit{functional causal models}, such as LiNGAM \citep{shimizu_2006} and Additive Noise Models \citep{hoyer_2008_ident}, directly output DAGs. Recent advancements falling in either of the classes above have considered methods relying on deep learning techniques \citep{zheng_2018, yu_2019}, but are specifically suited for large sample regimes.

MCMC-based Bayesian approaches, that account for graph uncertainty by learning a posterior distribution over graphs, include \cite{madigan1995bayesian} and \cite{giudici2003improving}, who introduced the MCMC Model Composition ($\text{MC}^3$) scheme and the Add-Delete-Reverse (ADR) scheme respectively. \cite{grzegorczyk2008improving} improve ADR's mixing by proposing a new arc reversal move. Another stream of contributions on structure learning MCMC have considered working with score functions in smaller spaces rather than the DAG one. \cite{friedman_2003} first recasted the problem by operating in the space of ``orders" instead of DAGs, so that candidate orders, consistent with multiple DAGs, are proposed. Although this generates considerable benefits in terms of scalability, it introduces sampling bias, as some DAGs are over-represented \citep{ellis2008learning}. This bias is addressed in \cite{kuipers_2017}, who proposed working in the space of ``ordered partitions" instead, which introduces scalability issues, making partition MCMC applicable to graphs with a small number of nodes. Further works have focused on improving mixing efficiency and scalability of these score-based MCMC \citep{niinimaki_2016_partialorder, vinikka_2020, kuipers_2022}. Recent advances have also considered approximate Variational Inference methods \citep{lorch_2021_DIBS, cundy2021bcd,geffner2022deep}. More closely related to our work, \cite{van2022g} considered the locally-informed proposal on non-decomposable Gaussian Graphical Models (GGM), but with neighborhoods defined by a subset of the space around the current state of the Markov chain, \textit{i.e.}~neighborhoods of the $\text{MC}^3$ type, with size growing quadratically in the number of nodes. However, their WWA algorithm does not target the posterior on DAGs directly, unlike PARNI-DAG.

Tangential to this work is also the rich literature on discrete spaces MCMC samplers for Bayesian variable selection. Noteworthy contributions include \cite{zanella2020informed}, that up-weights a random-walk kernel via local neighborhood information and constructs the \textit{locally-informed proposal}, and \cite{griffin2021search}, that develops an adaptive MCMC algorithm addressing high-dimensional sparse settings. The locally-informed proposal can be viewed as a discrete analog to the gradient MCMC proposals (\textit{e.g.}~the Metropolis-adjusted Langevin algorithm \citep{roberts1998optimal}), where the gradient of the target distribution does not exist. The adaptive MCMC instead is capable of learning from the shape of the target distribution and exploiting the most important variables. \cite{liang2022adaptive} generalise the locally-informed proposal and the adaptive MCMC to be used within random neighborhood schemes, and introduced the PARNI proposal for Bayesian variable selection.

\paragraph{Contribution.} This work presents a novel MCMC sampler for structure learning, \textbf{PARNI-DAG}, that builds on top of PARNI \citep{liang2022adaptive}, but is equipped with new features for efficient sampling in the space of DAGs. PARNI-DAG constructs a random neighborhood of possible DAGs via probabilities proportional to a function of the Posterior Edge Probabilities (PEPs), that is, the probabilities of including a direct edge between two nodes. Then, since full enumeration of all the possible DAGs within the neighborhood is virtually impossible in high-dimensions, it proposes a new candidate DAG via a ``point-wise update", \textit{i.e.}~by considering a sequence of intermediate proposed DAGs belonging to a smaller subset of the neighborhood. Finally, to guarantee better scalability with the number of nodes, we propose a procedure to pre-tune PARNI-DAG's parameters and warm-start the chain by utilising a skeleton graph derived through any constraint or score based algorithm (\textit{e.g.}~PC, GES, ...), or the iterative search space expansion proposed in \cite{kuipers_2022}.


\paragraph{Motivations.} As we will discuss in later sections, PARNI-DAG generates clear advantages over structure MCMC methods such as MC$^3$ and ADR in terms of speed of convergence and mixing, thanks to its new improved proposal, while targeting the same posterior distribution over DAGs. As for the comparison to score-based MCMC methods, it is known that order MCMC \citep{friedman_2003} has better scalability than standard structure MCMC methods, but, operating in the smaller space of order, it introduces sampling bias via a non-uniform prior being assigned to the different DAGs compatible with a single order \citep{ellis2008learning}. Compared to order MCMC, PARNI-DAG does not incur in any sampling bias as it targets the posterior on DAGs directly, while at the same time converging faster than ADR. \cite{kuipers_2017} propose a variant called partition MCMC, which operates in the space of ordered partitions instead of orders. Partition MCMC is unbiased in terms of sampling, but it is extremely slow and has high computational complexity (can only be used on very few nodes) \citep{kuipers_2022}. PARNI-DAG, on the other hand, specifically addresses high-dimensional settings with many nodes.

As a solution to scale order and partition MCMC up to high-dimensional scenarios, the hybrid MCMC scheme of \cite{kuipers_2022} proposes to restrict the initial search space by pruning it via a skeleton graph (\textit{e.g.}~PC algorithm derived skeleton) $\mathcal{H}$, whose max parent set size is $m$ per node, so that complexity reduces to $O(n 2^m)$. Although this increases MCMC speed dramatically in large graphs, it is likely to introduce bias as some true edges might be deleted in $\mathcal{H}$. To tackle this issue they also propose an iterative procedure (Iterative MCMC) to re-populate $\mathcal{H}$ with additional potential parent nodes, at the expenses of an increased computational time. We couple PARNI-DAG with a similar procedure, but we use a previously derived skeleton $\mathcal{H}$ (e.g.~the Iterative MCMC one) not to restrict the space, and possibly cancel out relevant edges, but to warm-start the chain by pre-tuning some of the MCMC, as we will explain in details in later sections.

We show in the experiments section how PARNI-DAG brings about improvements over the classical structure MCMC ADR \citep{giudici2003improving, grzegorczyk2008improving} and score-based MCMC methods \citep{friedman_2003, kuipers_2017, kuipers_2022} in terms of DAG learning accuracy and MCMC mixing, particularly in settings with a high number of nodes, as it is able to reach high probability regions in very few steps, while addressing the presence of highly correlated neighborhoods of edges where classical samplers might get trapped. Code to implement PARNI-DAG and fully reproduce the experiments is provided\footnote{\textbf{Code is provided in the supplementary materials. Github link to be added upon acceptance.}}.







\section{Problem Setup} \label{sec2:probsetup}

Consider a graph $\mathcal{G} = (\mathcal{V}, \mathcal{E})$, made of nodes $\mathcal{V}$ and edges $\mathcal{E}$, suitable to represent probabilistic dependencies (edges) between random variables (nodes) through a probabilistic graphical model \citep{koller2009probabilistic}. Given a collection of $n = | \mathcal{V} |$ continuous random variables $(X_1, ..., X_n)$, a Bayesian Network $\mathcal{B} = (\mathcal{G}, \Phi)$ associated with a DAG $\mathcal{G}$, is a probabilistic graphical model utilized to represent a factorization of the joint probability $p(X_1, ..., X_n)$ into the conditional distributions $ p(X_1, ..., X_n) = \prod^n_{i=1} p_{\phi} \big( X_j \, | \, \text{Pa}(X_j) \big) $, where $\phi \in \Phi$ are the parameters and $\text{Pa}(X_j)$ all the parents of node $X_j$. Suppose that all nodes $(X_1, ..., X_n)$ are re-scaled to have zero mean and unit variance. The goal in structure learning is to learn the BN factorization of conditional probabilities, given a sample of $N$ observations on $\mathcal{D} = \{X_{1,i}, ..., X_{n,i} \}^N_{i = 1}$. In Bayesian structure learning, we specifically want to learn a posterior distribution on possible DAGs, i.e.~$p(\mathcal{G} | \mathcal{D}) \propto p(\mathcal{D} | \mathcal{G}) p(\mathcal{G})$. In this work, we define a DAG indicator variable $\gamma \in \Gamma = \left\{0,1\right\}^{n \times n}$, where each $\gamma$ implies a unique graph $\mathcal{G}_\gamma$; $\gamma_{ij} = 1$ indicates the presence of an edge from $X_i$ to $X_j$, while $\gamma_{ij} = 0$ indicates absence of it. The linear functional model associated with DAG $\mathcal{G}_\gamma$ can be specified as
\begin{align}
    X =  W_\gamma ^{\top} X + \bm{\varepsilon}, \quad \text{ where} ~~ \mathbb{E} ( \bm{\varepsilon} ) = \bm{0}, ~~ \text{Var} (\bm{\varepsilon}) = \text{diag} ( \sigma^2_1, ..., \sigma^2_n) ~ ,  \label{eq:linFCM}
\end{align}
and $W_\gamma$ is a matrix of weights. Within a fully Bayesian approach to the inference problem presented in (\ref{eq:linFCM}), we consider the following prior specification 
\begin{align*} 
    & (W_\gamma)_{ij}| ~ \sigma_j^2, \gamma_{ij} = 1 ~ \sim ~ \text{Normal} (0, g\sigma_j^2)~, \\ & (W_\gamma)_{ij}| ~ \sigma_j^2, \gamma_{ij} = 0 ~ \sim ~ \delta_0 ~, \qquad p(\sigma_j) \propto \sigma_j^{-2}
\end{align*}
where $\delta_0$ is the Dirac measure at 0. Thanks to their conjugate form, the coefficients $\{W_{ij}\}_{i,j}$ and the heteroskedastic error variances $\{\sigma_j\}_{j}$ can be integrated out analytically, leading to a marginal likelihood that depends only on $\gamma$, denoted by $p(\mathcal{D}|\gamma)$. The last building block we need for posterior $\pi (\gamma) \equiv p(\gamma | \mathcal{D})$ inference is acyclicity constraints imposed in the prior distribution for the model indicator $\gamma$, which is specified by 
\begin{align}
    p(\gamma) \propto  \Big(\frac{h}{1-h}\Big)^{d_\gamma} \times ~ \mathbb{I} \big\{\text{$\mathcal{G}_\gamma$ is a DAG} \big\} \label{mat:prior}
\end{align}
where the hyperparameter $h \in (0,1)$ represents the prior edge probability and $d_\gamma = |\mathcal{E}_{\gamma}|$ is the number of edges implied by $\gamma$. We are interested in Bayesian inference on $\gamma$ with posterior distribution $\pi(\gamma) \propto p(\mathcal{D}|\gamma)p(\gamma)$. 

Notice that the likelihood, and thus posterior distribution, could also be replaced by a scoring function typically used in score-based MCMC methods \citep{friedman_2003, kuipers_2017, kuipers_2022}, such as the BGe score \citep{geiger2002parameter}, or the BDe score, which would make PARNI-DAG suitable also for discrete-valued nodes.


\section{The novel PARNI-DAG proposal}
\label{sec3:PARNI}

\subsection{Point-wise implementation of Adaptive Random neighborhood Informed proposal}

Recently, \cite{liang2022adaptive} introduced the Point-wise Adaptive Random Neighborhood Informed proposal (PARNI) in the context of Bayesian variable selection problems. PARNI is characterized by a \textit{Metropolis-Hastings} (MH) proposal \citep{metropolis1953equation,hastings1970monte} which couples the Adaptively scaled Individual Adaption scheme \citep{griffin2021search} with the locally-informed proposal \citep{zanella2020informed}. We begin by firstly describing the generic PARNI proposal for DAG learning problems, then we will introduce the modifications needed to make it efficient and that result in the PARNI-DAG proposal.

The PARNI proposal falls into the class of random neighborhood informed proposals, which are characterized by two steps: i) random sampling of a neighborhood $\mathcal{N}$, and then ii) proposal of a new DAG within this neighborhood $\mathcal{N}$ according to a informed proposal \citep{zanella2020informed}. The random neighborhoods are drawn based on an auxiliary \emph{neighborhood indicator} variable $k \in \mathcal{K}$, with conditional distribution $p(k|\gamma)$, such that the neighborhood is constructed as a function of $\mathcal{N}(\gamma, k) \subseteq \Gamma$. Let $\mathcal{K} = \Gamma = \{0,1\}^{n \times n}$, then in PARNI each $k \in \mathcal{K}$ value indicates whether the corresponding position of $\gamma$ is included in the neighborhood. 
The conditional distribution of $k$ takes a product form $p_{\eta}(k|\gamma) = \prod_{i,j} p_{\eta}(k_{ij}|\gamma_{ij})$, characterized by the set of tuning parameters $\eta = \{\eta_{ij}\}_{i,j=1}^n$, where $\eta_{ij} \in (\epsilon, 1-\epsilon)$ for a small $\epsilon \in (0, 1/2)$, and each $p_{\eta}(k_{ij}|\gamma_{ij})$ is given by
\begin{align} 
    p_{\eta}(k_{ij} = 1|\gamma_{ij} = 0) = \min\left\{1,\frac{1 - \eta_{ij}}{\eta_{ij}}\right\}~, \quad  p_{\eta}(k_{ij} = 1|\gamma_{ij} = 1) = \min\left\{1,\frac{\eta_{ij}}{1-\eta_{ij}}\right\}~.  \label{mat:PARNI_RN}
\end{align}
The methods for adapting $\eta$ will be discussed in Section \ref{subset:eta_tuning} specifically. The neighborhood is thus the constructed as
\begin{align}
    \mathcal{N}(\gamma, k) = \left\{\gamma^* \in \Gamma \mid \, \gamma^*_{ij} = \gamma_{ij} ~~ \forall (i,j) ~~ \text{s.t.} ~~ k_{ij} = 0 \right\}.
\end{align}
The neighborhood $\mathcal{N}(\gamma, k)$ contains $2^{d_k}$ models where $d_k$ denotes the number of 1s in $k$. The full enumeration over the whole neighborhood is computationally expensive if $d_k$ is big, and infeasible when $d_k$ is beyond 30. \cite{liang2022adaptive} considered a point-wise implementation of the algorithm, which dramatically reduces the computational complexity from $\mathcal{O}(2^{d_k})$ to $\mathcal{O}(2 d_k)$. The idea of how the point-wise implementation works is the following. Let variable $K = \{K_r\}_{r=1}^R$ represent the collection of positions where $k_{ij} = 1$. Instead of working with the full neighborhood $\mathcal{N}(\gamma, k)$, we construct a sequence of smaller neighborhoods $\{\mathcal{N}(\gamma(r), K_r)\}_{r=1}^R \subset \mathcal{N}(\gamma, k)$, and a new DAG $\gamma^\prime$ drawn from these neighborhoods $\{\mathcal{N}(\gamma(r), K_r)\}_{r=1}^R$ is sequentially proposed according to the sub-proposals $q_{g, K_r}(\gamma(r-1), \cdot)$ at each time $r$. Given the intermediate DAGs $\gamma = \gamma(0) \to \gamma(1) \to \dots \to \gamma(R) = \gamma^\prime$, each of their corresponding sub-proposal is defined by
\begin{align} 
    q_{g, K_r}(\gamma(r-1), \gamma(r)) = \frac{g\left(\frac{\pi(\gamma(r))p(K_r|\gamma(r))}{\pi(\gamma(r-1)p(K_r|\gamma(r-1)))}\right) \mathbb{I} \{\gamma(r) \in \mathcal{N}(\gamma(r-1), K_r)\}}{Z_r} ~ . \label{mat:within_PARNI}
\end{align}
where $g:[0, \infty) \to [0, \infty)$ is a continuous function and $Z_r$ is the normalising constant. The construction of $K$ and of the sub-neighborhoods $\{\mathcal{N}(\gamma(r), K_r)\}_{r=1}^R$ will be discussed later in Section \ref{subset:neighborhood}.

The choice of function $g(\cdot)$ is crucial to the performance of the PARNI proposal. If $g(x) = 1$, the sub-proposal $q_{g, K_r}$ in (\ref{mat:within_PARNI}) reverts back to a random walk proposal which proposes a new DAG from $N(\gamma(r-1), K_r)$ with uniform probability. It would then be hard for it to explore important neighborhoods of edges in high-dimensions, as it is likely to get stuck in neighborhoods with highly correlated nodes. In the locally-informed proposal of \cite{zanella2020informed}, function $g(\cdot)$ is chosen to be a non-decreasing and continuous balancing function $g:[0, \infty) \to [0, \infty)$, which satisfies $g(x) = xg(1/x)$. \cite{zanella2020informed} shows that the locally-informed proposal with balancing function $g(\cdot)$ is asymptotically optimal compared to other choices of $g(\cdot)$ in terms of Peskun ordering. In this work, we consider the Hastings' choice (\textit{i.e.} $g(x) = \min\{1,x\}$, satisfying $g(x) = xg(1/x)$) as it has been shown to be the most stable choice for most problems (see the discussion in Supplement B.1.3 of \cite{zanella2020informed}). 

The informed proposal in this context not only helps speed up convergence, but it also avoids proposing invalid DAGs. This is because the acyclicity checks for candidate models $\gamma^\prime \in \mathcal{N}(\gamma(r-1), K_r)$ can be run before computing their posterior model probabilities, and thus one can assign zero mass to their prior (resulting in zero mass posterior) if $\gamma^\prime$ is not a valid DAG. The resulting proposal kernel is then given by
\begin{align} 
    q_{g, K}(\gamma, \gamma^\prime) = \prod_{r = 1}^R q_{g, K_r} (\gamma(r-1), \gamma(r)).
\end{align}
In order to construct a $\pi$-reversible chain and calculate the MH acceptance probability, we define the new collection of variables $K^\prime = \{K_r^\prime\}_{r=1}^R$ for the reverse moves, where $K^\prime$ contains the same element as $K$ but with reverse order (\textit{i.e.} $K_r^\prime = K_{R-r+1}$). The MH acceptance probability is then defined as
\begin{align} 
    \alpha(\gamma, \gamma^\prime) = \min \left\{1 , \frac{\pi(\gamma^\prime) q_{g, K^\prime} (\gamma^\prime, \gamma)}{\pi(\gamma)q_{g, K} (\gamma, \gamma^\prime)} \right\}. \label{mat:MH_accprob}
\end{align}
\textbf{Proposition 3} of \cite{liang2022adaptive} shows that using the above $K^\prime$ can simplify the calculation of the MH acceptance probability in (\ref{mat:MH_accprob}).

In the following sections, we will introduce the necessary modifications and developments to efficiently sample DAGs, the combination of which defines the novel PARNI-DAG proposal. The full details and algorithmic pseudo code of PARNI-DAG are also provided in the supplementary material (Appendix D).

\subsection{Warm-start of $\eta$ combined with hybrid scheme}
\label{subset:eta_tuning}

\cite{griffin2021search} and \cite{liang2022adaptive} both suggested using the Posterior Inclusion Probabilities in a Bayesian variable selection context to define the $\eta_{ij}$ hyperparameters described above. Perfectly analogous, we will make use of the Posterior Edge Probabilities (PEPs) instead, \textit{i.e.}~$\pi(\gamma_{ij} = 1)$. The distribution in (\ref{mat:PARNI_RN}) with the choice of parameters described, encourages to `flip' edge values based on variables' marginal importance. For example, suppose the posterior for $\gamma_{ij}$ is $\pi(\gamma_{ij} = 1) = 0.9$, and the chain is currently at $\gamma_{ij} = 0$. The edge should be then assigned higher probability to be included in the DAG (\textit{i.e.}~$\gamma_{ij}$ should be switched from 0 to 1). Thus, we propose $k_{ij} = 1$ with probability $\min\{1, 0.9/0.1\} = 1$, so that the new neighborhood features this possible move and $\gamma_{ij}$ can be effectively switched from 0 to 1 in the next moves. Another advantage of using the conditional distribution featuring the PEPs in (\ref{mat:PARNI_RN}) is that when all variables are independent this proposal will have acceptance probability 1 and achieve optimal asymptotic variance as showed in \cite{griffin2021search}.

The PEPs, $\pi(\gamma_{ij} = 1)$, however cannot be directly calculated for most problems. \cite{griffin2021search} considered the Rao-Blackwellised estimates $\hat{\pi}(\gamma_{ij} = 1|\gamma_{-ij})$, where $\gamma_{-ij}$ denote $\gamma$ excluding the edge indicator $\gamma_{ij}$, and $\eta$ is updated on the fly given the current output of the chain. The Rao-Blackwellisation is however infeasible in the context of DAG sampling because it entails computing $n^2$ model probabilities at each iteration of the MCMC. This would result in an impractical computational scheme when working with many nodes. Alternatively, one could take the ergodic sum over the output $\{\gamma^{(l)}\}_{l=1}^t$ as
\begin{align}
    \hat{\pi}^{(t)}_{ij} = \frac{1}{t} \sum_{l = 1}^t \mathbb{I} \left\{ \gamma_{ij}^{(l)} = 1 \right\} ,
\end{align}
but the ergodic sum would be too slow to converge. This updating scheme would also lead to other issues such as poor exploration and mis-specification of the important edges. It is worth mentioning that the $\eta$'s do not have to be the exact PEPs, as we can still draw valid samples from the target posterior distribution $\pi$ as long as the MH acceptance probability preserves $\pi$-reversibility. However, choosing $\eta$'s that are close to the true PEPs will cause each component to change and to target the right proportion of their marginal probabilities (higher probability of including important edges).

Our solution then is to approximate the PEPs $\Tilde{\pi}_{ij}$ before actually running the PARNI-DAG's MCMC, and use this approximation to warm-start the chain. While running the algorithm at time $t$, we update the tuning parameters $\hat{\eta}^{(t)}_{ij}$ adaptively based on
\begin{align}
    \hat{\eta}^{(t)}_{ij} = \phi_t \Tilde{\pi}_{ij} + (1-\phi_t) \hat{\pi}^{(t)}_{ij} ~ ,
\end{align}
where $\{\phi_l\}_{l=1}^t$ is a decreasing sequence of weights which control the trade-off between the provided warm-start PEPs $\Tilde{\pi}_{ij}$ and the ergodic sum of the output $\hat{\pi}^t_{ij}$. The choice of $\{\phi_l\}_{l=1}^t$ is not unique and we give a full specification of our choice in the supplementary material (Appendix B). In what follows, we will briefly describe the methods used to calculate the warm-start approximation $\Tilde{\pi}_{ij}$.

We argue that the iterative procedure used to restrict the initial search space in \cite{kuipers_2022} can be particularly useful in our case to efficiently approximate the PEPs before running the chain. Following the same notation as in \cite{kuipers_2022}, let $\mathcal{H}$ be the adjacency matrix underlying a skeleton graph obtained from any DAG learning algorithm (\textit{e.g.}~PC, GES, etc.). For node $X_j$, the adjacency matrix $H$ defines the \textit{permissible parent set} $h^j = \left\{X_{i}: ~ H_{ij} = 1\right\}$. All possible parent combinations of node $X_j$ are included in the \textit{collection of permissible parent set} which is defined by
\begin{align}
    \mathbf{h}^j = \{ m |~ m \subseteq h^j \}.
\end{align}
The collection of permissible parent set $\mathbf{h}^j$ can be then extended to a bigger collection $\mathbf{h}^j_+$ as described in Section 4.3 of \cite{kuipers_2022}. The extended collection of permissible parent set $\mathbf{h}^j_+$ essentially includes one additional node from outside the permissible parent set $h^j$ as an additional parent, otherwise the candidate parents of $X_j$ are restricted to the set $h^j$. After deriving $\mathbf{h}^j_+$ via this procedure as in \cite{kuipers_2022}, we then approximate $\Tilde{\pi}_{ij}$ through the following procedure:
\begin{enumerate}[leftmargin=4ex]
    \item We remove the acyclicity constraint from the prior (\ref{mat:prior}) and compute the unconstrained posterior distribution $\pi^u(\gamma)$, approximating the marginal probability $\pi^u(\gamma_{ij})$ under $\mathbf{h}^j_+$.
    \item We approximate the joint probability of a pair edges by the product
    \begin{align}
        \Tilde{\pi}(\gamma_{ij}, \gamma_{ji}) \approx \pi^u(\gamma_{ij}) \times \pi^u(\gamma_{ji}).
    \end{align}
    \item We impose the simplest acyclicity constraint which avoids creating two edges with opposite directions, \textit{i.e.} let $\Tilde{\pi}(\gamma_{ij} = 1, \gamma_{ji} = 1) = 0$. Approximate 
    \begin{align}
        \Tilde{\pi}_{ij} = \frac{\Tilde{\pi}(\gamma_{ij} = 1, \gamma_{ji} = 0)}{\Tilde{\pi}(\gamma_{ij} = 1, \gamma_{ji} = 0) + \Tilde{\pi}(\gamma_{ij} = 0, \gamma_{ji} = 1) + \Tilde{\pi}(\gamma_{ij} = 0, \gamma_{ji} = 0)}.
    \end{align}
\end{enumerate}
The full details for the calculations of $\Tilde{\pi}_{ij}$ are given in the supplementary material (Appendix C).

\subsection{Introducing the ``reversal'' neighborhood}
\label{subset:neighborhood}

In the original PARNI proposal for Bayesian variable selection, each $K_r$ in the intermediate proposals corresponds to exactly one position $(i,j)$, such that $k_{ij} = 1$ and that each intermediate neighborhood $\mathcal{N} (\gamma(r-1), K_r)$ only contains two DAGs: $\gamma(r-1)$ itself and the DAG obtained by flipping $K_r$ in $\gamma(r-1)$. This sequential neighborhood construction is clearly not appropriate for the problem of DAG sampling and might result in very slow mixing. In addition, it is natural to assume that a pair of edges $i \longleftrightarrow j$ are highly correlated and thus equally likely to be included in a DAG. It is then necessary to specify a reversal move that achieves good mixing between pairs of edges. The neighborhood construction in the original PARNI proposal does not support a reversal move in one single sub-proposal. To elaborate on this, suppose that currently $\gamma_{ij} = 1$ and $\gamma_{ji} = 0$, to propose a reversal move according to the original PARNI proposal, we should first propose to remove $\gamma_{ij}$ from the DAG, then include $\gamma_{ji}$ in the following sub-proposals. These flips might be running into issues if the posterior model probability for excluding both edges, $\gamma_{ij} = \gamma_{ji} = 0$, is low, so that these moves are rarely proposed and the chain gets stuck at $\gamma_{ij} = 1$. 

Therefore, we construct a bigger neighborhood containing the edge reversal move of a DAG. This modification allows for an edge reversal move within a single sub-proposal $q_{g, K_r}$ without going through the low posterior probability models and avoids the chain getting stuck. To clarify this, suppose $k$ is sampled at each iteration of the chain. For all those pairs which satisfy $k_{ij} = 1$ and $k_{ji} = 1$ in $k$, we construct the reversal neighborhoods for edges $\gamma_{ij}$ and $\gamma_{ji}$. For other positions such that $k_{ij} = 1$, but $k_{ji} = 0$, we construct the original neighborhood which only contains the current model and model with $\gamma_{ij}$ flipped. This reversal neighborhood specification is included into the main PARNI-DAG proposal given in the supplementary material (Appendix D). 

\subsection{New adaptive scheme for better computational efficiency}

Although the point-wise implementation reduces the per-iteration costs from exponential to linear in the neighborhood size (equivalent to the number of sub-neighborhoods in PARNI), the PARNI proposal is still computationally expensive when the sampled DAGs are not very sparse. For example, considering the \texttt{gsim100} simulated data of \cite{suter2021bayesian}, featuring 161 true edges, we would be often required to compute 320 posterior DAG probabilities on average in a single MCMC iteration, and the computation thus becomes 320 times more expensive than the standard ADR proposal in this case. To reduce the computational cost of the PARNI-DAG proposal even more, we make use of a `neighborhood thinning' parameter $\omega$ to control the number of neighborhoods to be evaluated.

In the original PARNI proposal, the thinning parameter $\omega$ represents the random walk jumping probability, so it does not take the posterior inclusion probabilities into account and does not affect the number of neighborhoods evaluated. For the new PARNI-DAG proposal instead, with probability $1-\omega$, we will skip evaluation of some of the neighborhoods. Within each iteration, for each point $r$ in the sequence, given the current sub-proposal $\gamma(r-1)$ and sub-neighborhood indicator $K_r$, PARNI-DAG:
\begin{enumerate}[leftmargin=5ex]
    \item with probability $\omega$, proposes $\gamma(r) \sim q_{K_r}(\gamma(r-1), \cdot)$ as in (\ref{mat:within_PARNI}).
    \item with probability $1-\omega$, does not evaluate the posterior probabilities of other DAGs in $\mathcal{N}(\gamma(r-1), K_r)$, and directly propose $\gamma(r) = \gamma(r-1)$.
\end{enumerate}
Varying $\omega$ can vary the number of posterior DAG probability evaluations. The larger the value of $\omega$, the more the DAG probabilities evaluated. If $\omega = 1$, all neighborhoods are required to be evaluated to propose a new DAG. 

The parameter $\omega$ is updated adaptively. Here, we consider the Robins-Monro adaptation scheme to update $\omega$ on the fly. At time $t$, the new $\omega^{(t+1)}$ parameter is updated via
\begin{align}
    \text{logit}_\epsilon \omega^{(t+1)} = \text{logit}_\epsilon \omega^{(t)} - \psi_t (\mathcal{N}_t - \Tilde{\mathcal{N}})
\end{align}
where $\mathcal{N}_t$ is number of sub-neighborhoods evaluated at time $t$ and $\Tilde{\mathcal{N}}$ is a pre-specified expected number of evaluations. We set $\psi_t = t^{-0.7}$, meaning that the adaptation of $\omega$ diminishes at the rate of $\mathcal{O}(t^{-0.7})$. From the empirical results, we find that $\Tilde{\mathcal{N}} = 10$ evaluated neighborhoods on average seems to achieve optimal balance between mixing performance and computational cost. Therefore, we choose to set $\Tilde{\mathcal{N}} = 10$ for most of the numerical studies in the next section and in the supplementary material.

\section{Experiments} \label{sec5:simulations}

\subsection{Convergence and mixing efficiency}

\begin{table*}[t]
    \begin{minipage}[h]{0.70\textwidth}
    \centering \includegraphics[width=\textwidth]{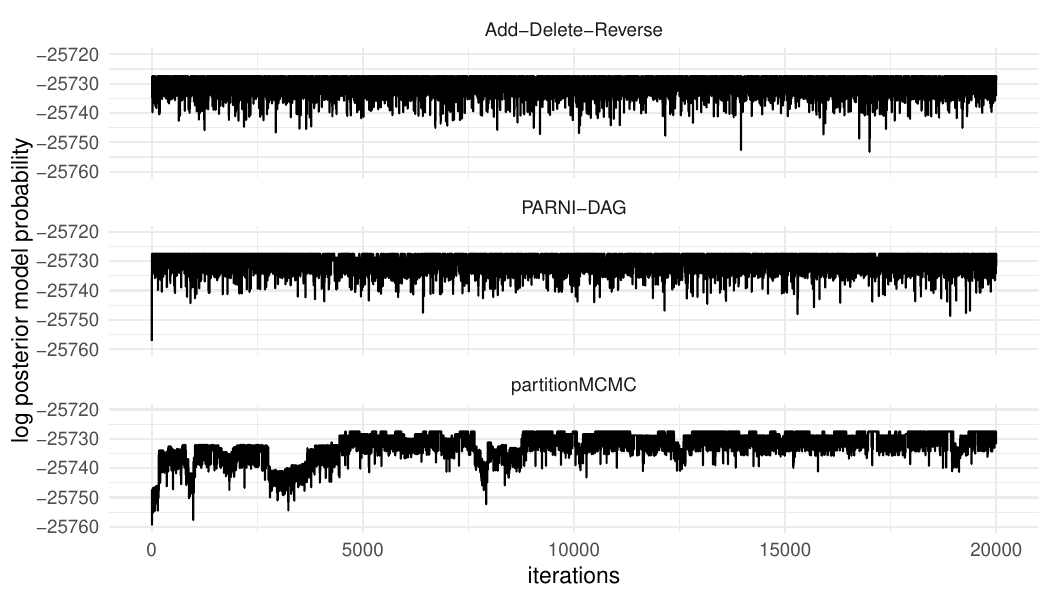}
    \captionof{figure}{Protein dataset: trace plots of log posterior model probabilities. All algorithms were ran for the same CPU-time and thinned to 20,000 measures.}
    \label{fig1:protein}
  \end{minipage}
  \hfill
  \begin{minipage}[h]{0.28\textwidth}
  \ra{1.2}
    \centering
    \begin{tabular}{c|c}
\toprule
algorithm & MSE \\
\midrule 
\textbf{ADR}    & 8.84 \\
\textbf{PARNI-DAG} &  19.44 \\
\textbf{Order MCMC}  &  145.17  \\
 \textbf{Partition MCMC} &  121.83  \\
\bottomrule
\end{tabular}
      \captionof{table}{Protein dataset: time-normalised median MSE ($\times 10^{-8}$) on estimating posterior edge probabilities. (\textit{Lower is better.})}
      \label{tab1:protein}
    \end{minipage}
\end{table*}

\paragraph{Protein data.} We first consider the real-world protein-signalling dataset \citep{sachs2005causal}, found also in \cite{cundy2021bcd}, to test PARNI-DAG's mixing. The dataset consists of $n = 11$ nodes and $N = 853$ observations and the ground-truth DAG (17 edges) is provided by expert knowledge. We compare the performance of PARNI-DAG to ADR, Order MCMC and Partition MCMC. We provide the full details on the implementation of the compared models in the supplementary materials (Appendix E). For a fair comparison of the models, we re-define the score functions for Order and Partition MCMC such that all schemes target the same posterior distribution as specified in Section \ref{sec2:probsetup}. Notice that for this dataset, it is feasible for all the four MCMC schemes to use the full skeleton, without the need to restrict the initial space \citep{kuipers_2022}. For ADR and PARNI-DAG, we use prior parameters $g = 10$ and $h = 1/11$.

We compare trace plots of log posterior DAG probabilities for ADR, PARNI-DAG and Partition MCMC in Figure \ref{fig1:protein}. Since we could not extract the full score trace of Order MCMC using the \texttt{BiDAG} package \citep{suter2021bayesian}, Order MCMC is not included in Figure \ref{fig1:protein}. We ran ADR, PARNI-DAG and Partition MCMC for 480,000, 60,000 and 20,000 iterations respectively which take about the same CPU time\footnote{Using Intel i7 2.80 GHz processor}, and thin the output from ADR and PARNI-DAG so that 20,000 measures are kept. From Figure \ref{fig1:protein}, all these three algorithms mix equally well, the only difference is that the Partition MCMC algorithm explores more DAGs with lower probabilities. We then compare the median mean squared errors (MSE) on estimating PEPs compared to a ground-truth estimate. The MSE assesses both the convergence of the chain and the bias in the Monte Carlo estimates. The ground-truth PEPs were obtained by running Partition MCMC for approximately 5 hours. We ran all algorithms for 20 replications. Each individual chain was run for 3 minutes. The MSE estimates are presented in Table \ref{tab1:protein}. The structure MCMC schemes (ADR and PARNI-DAG) generally outperform Order MCMC and Partition MCMC and return estimates with lower MSE. Notice that ADR in this specific case also outperforms PARNI-DAG, since the dataset is very low-dimensional (11 nodes), with PARNI-DAG not trailing very far behind. We will demonstrate in the next experiments how PARNI-DAG instead performs better in higher-dimensional settings.

\paragraph{gsim100 data.} We also study the performance of ADR, PARNI-DAG, Order MCMC and Partition MCMC on a more complex graph, the \texttt{gsim100} simulated dataset, found in the \texttt{BiDAG} package \citep{suter2021bayesian}. This features a randomly generated DAG with $n = 100$ nodes and $N = 100$ observations, with 161 true edges. For ADR and PARNI-DAG we set prior parameters to $g = 10$ and $h = 1/100$. Since $n > 20$, it is not feasible to use the full skeleton as in the protein dataset. So we consider two skeletons to restrict the initial space: PC derived skeleton $\mathcal{H}_{\text{PC}}$ and the skeleton derived with the iterative procedure of \cite{kuipers_2022} $\mathcal{H}_{\text{iter}}$, featuring more edges than $\mathcal{H}_{\text{PC}}$.

We examine the trace plot of log posterior DAG probabilities from ADR, PARNI-DAG and Partition MCMC using both skeletons $\mathcal{H}_{\text{PC}}$ and $\mathcal{H}_{\text{iter}}$ respectively (Figure \ref{fig2:gsim100}). We ran ADR, PARNI-DAG and Partition MCMC for 80,000, 20,000 and 20,000 iterations respectively. Notice that, under both $\mathcal{H}_{\text{PC}}$ and $\mathcal{H}_{\text{iter}}$, PARNI-DAG converges very quickly to the high-probabilistic region, whereas ADR takes longer. Partition MCMC under $\mathcal{H}_{\text{PC}}$ cannot reach the same high-probability region as it targets a posterior distribution restricted to $\mathcal{H}_{\text{PC}}$, which is biased if $\mathcal{H}_{\text{PC}}$ excludes true positive edges. Under the more populated $\mathcal{H}_{\text{iter}}$ instead it converges, but still slower than PARNI-DAG. When comparing median MSEs on PEPs on this dataset, we considered 20 replications with each individual chain being run for 15 minutes. As for Order and Partition MCMC, $\mathcal{H}_{\text{PC}}$ seems to lead to slightly better performance than $\mathcal{H}_{\text{iter}}$. This might be due to the fact that in this high-dimensional example $\mathcal{H}_{\text{iter}}$ includes considerably more DAGs than $\mathcal{H}_{\text{PC}}$, so Order MCMC, and even more so Partition MCMC, take longer to explore the space. As for ADR and PARNI-DAG, the choice of $\mathcal{H}$ does not have a big impact, as in their case this is used only to warm-start the chain and not to restrict the space. PARNI-DAG here outperforms ADR and it is competitive to Order MCMC as well, without introducing biases stemming from search space restriction and sampling.


\begin{table*}[t]
    \begin{minipage}[h]{0.64\textwidth}
    \centering \includegraphics[width=\textwidth]{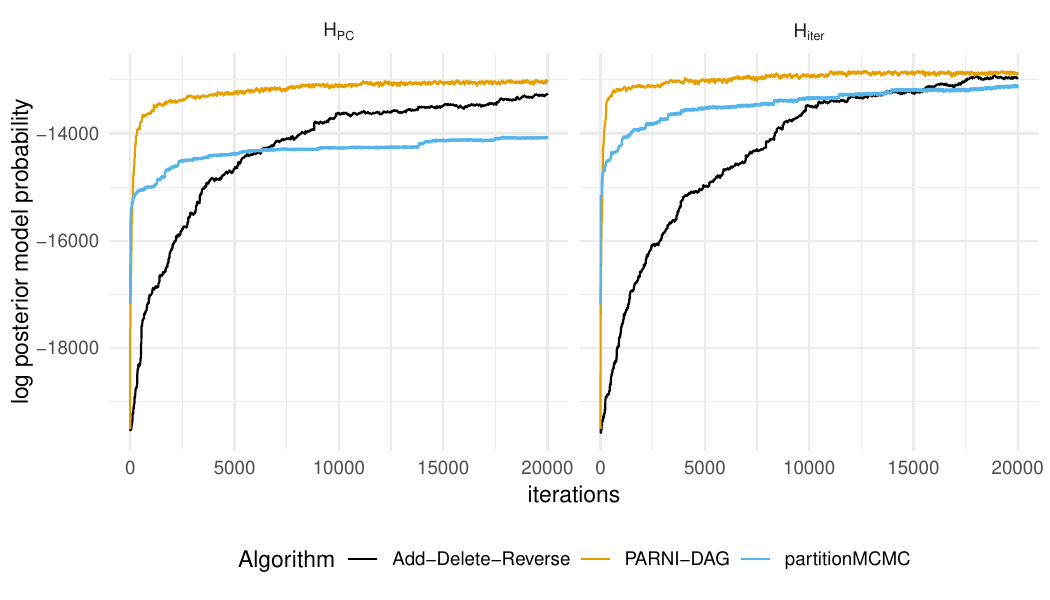}
    \captionof{figure}{\texttt{gsim100} dataset: trace plots of log posterior model probabilities. All algorithms were ran for the same CPU-time and thinned to 20,000 measures.} \label{fig2:gsim100}
  \end{minipage}
  \hfill
  \begin{minipage}[h]{0.34\textwidth}
  \ra{1.1}
    \centering
    \scalebox{0.8}{ 
    \begin{tabular}{c|c|c}
    \toprule
    skeleton     & algorithm      & MSE \\ 
    \midrule 
    \multirow{4}{*}{$\mathcal{H}_{\text{PC}}$}        & \textbf{ADR}           & 2.91        \\
                           & \textbf{PARNI-DAG}     & 1.83        \\
                           & \textbf{Order MCMC}   & 1.79    \\
                           & \textbf{Partition MCMC} & 12.75       \\ 
                           \midrule
    \multirow{4}{*}{$\mathcal{H}_{\text{iter}}$} & \textbf{ADR}           & 2.37        \\
                           & \textbf{PARNI-DAG}     & 1,72        \\
                           & \textbf{Order MCMC}   &   1.84  \\
                           & \textbf{Partition MCMC} & 31.16        \\ 
    \bottomrule
    \end{tabular}
    }
    \caption{table}{\texttt{gsim100} dataset: time-normalised median MSE ($\times 10^{-7}$) on estimating posterior edge probabilities. (\textit{Lower is better.})}
    \label{tab2:gsim_100}
    \end{minipage}
\end{table*}

\subsection{DAG learning accuracy}


Lastly, we compare performance of PARNI-DAG with other main methods in recovering the underlying true DAG on four different real-world graph structures. These graph structures are taken from the \texttt{R} package \texttt{bnlearn}\footnote{More information on: \url{http://bnlearn.com/}.} and include: i) \textit{ecoli70} dataset, with $n = 46$ nodes and $70$ edges; ii) \textit{magic-niab} dataset, with $n = 44$ nodes and $66$ edges; iii) \textit{magic-irri} dataset, with $n= 64$ nodes and $102$ edges; iv) \textit{arth150} dataset, with $n = 107$ nodes and $150$ edges. Given the graph structure and its associated parameters, we simulate $N=100$ \textit{i.i.d.}~observations for each of the four cases. The models we compare include: i) \textbf{PC} algorithm; ii) \textbf{GES}; iii) \textbf{LiNGAM}; iv) \textbf{DiBS}, a differentiable variational method introduced by \cite{lorch_2021_DIBS}; v) \textbf{DiBS$+$}, a version of DiBS with the weighted particle mixture; vi) \textbf{DECI}, another non-linear additive noise model that uses variational inference introduced by \cite{geffner2022deep}; vii) \textbf{Order MCMC}, on a restricted space defined by the skeleton of the PC algorithm; viii) \textbf{Iterative MCMC}, the version of Order MCMC run on a search space defined by starting with the PC skeleton, and iteratively expanding the possible parent sets until the score cannot be improved \citep{kuipers_2022}; ix) \textbf{Partition MCMC}, on the same space defined by the iterative procedure in Iterative MCMC; x) \textbf{ADR} scheme and xi) \textbf{PARNI-DAG}, both with pre-tuning of the $\eta$'s using the Iterative-derived skeleton. All the score-based methods (GES, Order, Partition, Iterative) are assigned an edge-penalty parameter equal to $2 log(n)$, as suggested in \citep{kuipers_2022}, to adjust to potentially sparse graphs. Performance is measured via the Structural Hamming Distance between the true DAG and the estimated DAG, averaged over $20$ replications of the experiment, for each dataset. Results are collected in Table \ref{tab:bnlearn}. Notice how PARNI-DAG is either the best or the second best method across all four experiments. In particular, it statistically matches the performance of the resulting best MCMC scheme in the first three experiments featuring a lower number of nodes, while it significantly outperforms all the methods in the case with high number of nodes (\textit{arth150}), as expected. All variational methods (DiBS, DiBS$+$ and DECI) struggle to detect edges in these low-sample setups as they require a stronger signal (or a larger sample, as featured in the experiments in \cite{lorch_2021_DIBS, geffner2022deep}) to perform well. Additionally, they are significantly slow to converge and run into numerical instability issues in the last large-$n$ dataset (\textit{arth150}).

\section{Limitations}
\label{sec:limitations}

Although the PARNI-DAG proposal has demonstrated significant improvements over the state-of-the-art algorithms considered in the previous section, it still comes with a few limitations that can be addressed. The first limitation lies in the linearity assumption. In fact, in order to derive the closed-form marginal likelihood featured in the locally informed proposal, PARNI-DAG is currently constrained to the linear model assumption described in (\ref{eq:linFCM}). This can potentially result in model misspecification in presence of non-linear relationships, which is a threat in cases where this hampers edges identifiability. The recent work of \cite{liang2023adaptive} has shed light on how to efficiently extend the PARNI proposal to more general non-linear models, and can possibly be adjusted to Bayesian structure learning settings that employ Additive Noise Models (ANMs) \citep{hoyer2008nonlinear}, where a closed-form marginal likelihood does not exist. The second limitation of the PARNI-DAG proposal is the computational cost associated with the likelihood evaluation, which scales at least linearly in the number of datapoints. The issue arises as the likelihood features both in the informed proposals and in the MH step. A viable solution to this problem consists in the possibility of coupling PARNI-DAG with optimal sub-sampling MCMC procedures, such as the ones presented in \cite{korattikara2014austerity} and \cite{maclaurin2014firefly} (FireflyMC). These limitations associated with the PARNI-DAG proposal provide us with interesting future research directions, aimed at improving its scalability and capability to handle more complex data.

\begin{table*}[t]
    \centering
    \ra{1.2}
    \scalebox{0.9}{
    \begin{tabular}{c|c|c|c|c}
    \toprule
            & \textbf{ecoli70} & \textbf{magic-niab} & \textbf{magic-irri} & \textbf{arth150}  \\
         \midrule
         \textbf{PC} & 62.65 $\pm$ 1.64 & 66.95 $\pm$ 0.82 & 103.75 $\pm$ 0.91 & 126.70 $\pm$ 2.10 \\
         \textbf{GES} & 47.95 $\pm$ 1.72 & 65.50 $\pm$ 0.42 & 97.60 $\pm$ 0.99 & 129.70 $\pm$ 3.47 \\
         \textbf{LiNGAM} & 107.4 $\pm$ 3.96 & 71.60 $\pm$ 1.34 & 111.25 $\pm$ 2.34 & -  \\
         \midrule
         \textbf{DiBS} & 71.05 $\pm$ 0.17 & 65.80 $\pm$ 0.24 & 102.35 $\pm$ 0.17 & - \\
         \textbf{DiBS$+$}  & 71.25 $\pm$ 0.28 & 64.70 $\pm$ 0.36 & 100.45 $\pm$ 0.40 & -  \\
         \textbf{DECI}  & 70.95 $\pm$ 0.02 & 66.95 $\pm$ 0.02  &  102.00 $\pm$ 0.00 &  -  \\
         \midrule
         \textbf{Order MCMC} & 42.85 $\pm$ 1.30 & \textbf{62.05 $\pm$ 1.52} & \textbf{92.90 $\pm$ 1.46} & 110.00 $\pm$ 1.96 \\
         \textbf{Iterative MCMC} & \textbf{39.55 $\pm$ 2.15} & 65.00 $\pm$ 2.43 & 96.15 $\pm$ 2.11 & 119.45 $\pm$ 3.43 \\
         \textbf{Partition MCMC} & 48.85 $\pm$ 2.76 & 66.20 $\pm$ 1.62 & 103.35 $\pm$ 1.81 & 140.75 $\pm$ 4.04 \\
         \textbf{ADR} & \textbf{39.40 $\pm$ 1.27} & 64.25 $\pm$ 0.64 & 95.20 $\pm$ 1.16 & 143.15 $\pm$ 1.62 \\
         \textbf{PARNI-DAG} & \textbf{37.55 $\pm$ 1.59} & \textbf{64.14 $\pm$ 0.55} & \textbf{91.85 $\pm$ 0.97} & \textbf{97.60 $\pm$ 2.16} \\
         \bottomrule
    \end{tabular}
    }
    \caption{SHD averaged over $20$ replications with associated 95\% confidence intervals of the various models compared on the four datasets ($n = 100$). All algorithms were ran for approximately the same CPU-time and the significantly best performing models are given in \textbf{bold}. (\textit{Lower is better.})}
    \label{tab:bnlearn}
\end{table*}

\section{Discussion}  \label{sec6:conclusions}

In this work, we proposed a novel MCMC sampler, PARNI-DAG, for a fully Bayesian approach to the problem of structure learning. PARNI-DAG samples directly from the space of DAGs and introduces improvements in terms of computational complexity and MCMC mixing, stemming from the nature of its adaptive random neighborhood informed proposal. This proposal facilitates moves to higher probability regions, and is particularly useful when dealing with high-dimensional settings with a high number of nodes, as demonstrated in the experimental section. In future work, we will consider adding global moves to the PARNI-DAG proposal, which re-generates a new DAG model every few iterations, to specifically tackle settings with highly correlated structures. 

In addition, we are also interested in studying PARNI's theoretical mixing time bounds. In fact, although the mixing time bound for random walk proposals has been largely studied, similar results for the class of locally informed proposals are relatively under-developed, due to the complexities arising in the proposal distributions. Most results on mixing time in discrete sample spaces focus on the problem of Bayesian variable selection. It has been shown by \cite{yang2016computational} that a random walk proposal (specifically the add-delete-swap proposal) can achieve polynomial-time mixing under mild conditions on the posterior distribution. Under similar conditions, \cite{zhou2022dimension} has shown that the mixing time of the Locally Informed and Thresholded proposal (LIT) does not depend on the number of covariates. The mixing time bound of MCMC samplers in Bayesian structure learning settings is notoriously harder to study, due to the higher complexity of the sample space of DAGs. The recent work of \cite{zhou2023complexity} has shed light on the analysis of mixing time bounds for Bayesian structure learning as a generalisation of the results in \cite{yang2016computational}. \cite{zhou2023complexity} has proven that the mixing time of the Random Walk Greedy Equivalent Search (RW-GES) proposal is at most linear in the number of covariates and the number of datapoints. Moreover, they also presented the necessary conditions for posterior consistency in Bayesian structure learning. It has not been formally proven yet that the informed proposal achieves faster theoretical mixing time compared to the random walk proposal in the context of Bayesian structure learning, although the empirical experiments we conducted suggest so. Similarly, it is highly likely then that the PARNI-DAG proposal can also achieve dimension-free mixing, such as the LIT proposal. We leave the topic of theoretical mixing time bounds of the PARNI proposal in different applications (e.g., Bayesian variable selection, Bayesian structure learning) to future research.


\medskip

\section*{Acknowledgements}

SL is supported by EPSRC research grant EP/V055380/1. AC's research is funded by the Defence Science and Technology Laboratory (Dstl) which is an executive agency of the UK Ministry of Defence providing world class expertise and delivering cutting-edge science and technology for the benefit of the nation and allies. The research supports the Autonomous Resilient Cyber Defence (ARCD) project within the Dstl Cyber Defence Enhancement programme.

\bibliography{Refs}


\end{document}


\maketitle

\appendix

\section{Derivation of the posterior model probability of DAG model}

In this section, we will introduce the companion form for the prior on $W_\gamma$ and derive the marginal likelihood of $\gamma$ starting from a spike-and-slab prior, as described in Section 2. In addition, under the \textit{Markov property}, we show that the posterior distribution $\pi (\cdot)$ can be factorised into a product form \citep{koller2009probabilistic}.

Let $\gamma_j$ be the $j$-th column of of $\gamma$, $\gamma_{ij}$ is the $i$-th component of $\gamma_j$ (equivalent to the $i,j$-th entry of indicator matrix $\gamma$). Let $\gamma_{-i,j}$ be $\gamma_j$ but excluding the i-th component $\gamma_{ij}$. Let $X_{\gamma_j}$ denote the ``active" variables in $\gamma_j$ (those variables such that $\gamma_{ij}$ = 1). The variables $X_{\gamma_j}$ are the parents of $X_j$ implied by $\gamma$. Let $W_{\gamma, j}$ be the coefficients of $j$-th column in $W_\gamma$ selected by $\gamma_j$. In addition, let $\bm{X}_i = \{X_{j,i}\}_{j=1}^n$ be a vector of nodes for a fixed $i$-th observation, with $X_j$ being the $j$-th node, while let $x_j = \{X_{j,i}\}_{i=1}^N$ be the vector of the $N$ observations corresponding to the $j$-th node. The matrix $x_{\gamma_j}$ is the collection of observations on the set of variables defined by $\gamma_j$.

Under $\gamma$, the parent set of $X_j$ ($\text{Pa}(X_j)$) is equivalent to set of nodes collected in $\gamma_j$ (i.e., $X_{\gamma_j}$). The joint prior distribution on $W_{\gamma}$ then is
\begin{align}
    p(W_\gamma|\gamma, \{\sigma_j^2\}) = \prod_{j = 1}^n   p(W_{\gamma, j}|\gamma_j, \sigma_j^2)
\end{align}
where
\begin{align}
    W_{\gamma, j}|~\gamma_j, \sigma^2_j \sim N_{q_j}(0, g \sigma_j^2 I_{q_j}) \label{mat:joint_prior}
\end{align}
and $q_j$ is the the parent size of $X_j$.

\begin{remark}
    The identity matrix $I_{q_j}$ in the prior (\ref{mat:joint_prior}) can be replaced by the inverse of the Gram matrix defined as $(x_{\gamma_j}^T x_{\gamma_j})^{-1}$. The resulting prior is then an analog, in the context of DAG sampling, to $g$-prior in the Bayesian variable selection context.
\end{remark}

Note that in a DAG model, each node $X_j$ is independent of its non-descendants given its parents $X_{\gamma_j}$. This property implies a factorisation of the joint likelihood over data $\mathcal{D}$ given by
\begin{align*}
    p(\mathcal{D}|W_\gamma, \{\sigma^2_j\}, \gamma) & = 
    \prod_{j=1}^n p(X_j|X_{\gamma_j}, W_{\gamma,j}, \sigma^2_j, \gamma_j),
\end{align*}
which translates into 
\begin{align*}
    p(\mathcal{D}|W_\gamma, \{\sigma^2_j\}, \gamma) & =  \prod_{j=1}^n N (x_j|x^T_{\gamma_j}W_{\gamma,j}, \sigma^2_j I_{N})
\end{align*}
in the context of Gaussian graphical models. Due to the factorisation above, it is possible to exchange the integral of $W_\gamma$ and $\{\sigma_j^2\}$ with the product in $j$, and integrate out $W_{\gamma,j}$ and $\sigma_j^2$ from the joint probabilities
\begin{align}
    p(X_j, W_{\gamma,j}, \sigma^2_j|X_{\gamma_j}, \gamma_j) = p(X_j|X_{\gamma_j}, W_{\gamma, j}, \sigma^2_j, \gamma_j) p(W_{\gamma,j}|\sigma^2_j, \gamma_j) p(\sigma^2_j)
\end{align}
for each individual node $X_j$. In what follows, we will show how to derive $p(X_j|X_{\gamma_j}, \gamma_j)$ from the above, by solving these integrals.

We start from the integration of the coefficients $W_{\gamma,j}$. We can marginalise out $W_{\gamma, j}$ simply due to conjugacy as below
\begin{align*}
    p(X_j|X_{\gamma_j}, \sigma^2_j, \gamma_j) & = \int p(X_j|X_{\gamma_j}, W_{\gamma, j}, \sigma^2_j, \gamma_j) p(W_{\gamma, j}|\sigma^2_j, \gamma_j) d W_{\gamma,j} \\
    & = \int N (x_j|x^T_{\gamma_j}W_{ \gamma,j}, \sigma^2_j I_{n}) N(W_{\gamma, j}| 0, g \sigma^2_j I_{q_j}) d W_{\gamma,j} \\
    & = (2\pi\sigma_j^2)^{-\frac{N}{2}} g^{-\frac{q_j}{2}} \det(\Sigma_j)^{-\frac{1}{2}} \exp\left\{-\frac{1}{2\sigma_j^2}(x_j^T x_j - x_j^T x_{\gamma_j}\Sigma_j^{-1}x^T_{\gamma_j}x_j)\right\}
\end{align*}
where $\Sigma_j = x^T_{\gamma_j}x_{\gamma_j} +\frac{1}{g}I_{q_j}$. Recall that the prior for $\sigma^2_j$ is specified by $p(\sigma_j^2) \propto \sigma_j^{-2}$, and thus we can marginalise out $\sigma_j^{2}$ in a similar way
\begin{align}
    p(X_j|X_{\gamma_j}, \gamma_j) & = \int p(X_j|X_{\gamma_j}, \sigma^2_j, \gamma_j) p(\sigma^2_j) d\sigma^2_j \notag \\
    & = \int_{\mathbb{R}^+} (2\pi\sigma_j^2)^{-\frac{N}{2}} g^{-\frac{q_j}{2}} \det(\Sigma_j)^{-\frac{1}{2}} \exp\left\{-\frac{1}{2\sigma_j^2}(x_j^T x_j - x_j^T x_{\gamma_j}\Sigma_j^{-1}x^T_{\gamma_j}x_j)\right\} \cdot \sigma_j^{-2} d\sigma^2_j \notag \\
    & = (2\pi)^{-\frac{N}{2}} g^{-\frac{q_j}{2}} \det(\Sigma_j)^{-\frac{1}{2}} \Gamma\left(\frac{N}{2}\right) \left[\frac{1}{2}(x_j^T x_j - x_j^T x_{\gamma_j}\Sigma_j^{-1}x^T_{\gamma_j}x_j)\right]^{-\frac{N}{2}}
\end{align}
where $\Sigma_j = x^T_{\gamma_j}x_{\gamma_j} +\frac{1}{g}I_{p_j}$ and $\Gamma(\cdot)$ is the Gamma function. Eventually, putting all the steps above together, we can write down the model's marginal likelihood, conditional on $\gamma$ only, as follows
\begin{align}
    p(\mathcal{D}|\gamma) & = \int\int p(\mathcal{D}|W_\gamma, \{\sigma^2_j\}, \gamma)dW_\gamma d \{\sigma^2_j\} \notag  \\
    & = \prod_{j = 1}^n \int \int p(X_j|X_{\gamma_j},  \beta_{j, \gamma_j}, \sigma_j^2) dW_{\gamma,j} d\sigma_j^2 \notag \\
    & = \prod_{j = 1}^n p(X_j|X_{\gamma_j},\gamma_j) \notag \\
    & \propto \prod_{j = 1}^n  g^{-\frac{q_j}{2}} \det(\Sigma_j)^{-\frac{1}{2}} \left[\frac{1}{2}(x_j^T x_j - x_j^T x_{\gamma_j}\Sigma_j^{-1}x^T_{\gamma_j}x_j)\right]^{-\frac{N}{2}}
\end{align}
where $\Sigma_j = x^T_{\gamma_j}x_{\gamma_j} +\frac{1}{g}I_{p_j}$. From the above derivation, we can also conclude the following property for the posterior distribution $\pi(\cdot)$:

\begin{property}[Markov property.] \label{propty:markov}
    The posterior distribution of indicator variable can be factorised as the following
    \begin{align}
        \pi(\gamma) \propto \prod_{j = 1}^n p(\gamma_j|\mathcal{D})  \times  \mathbb{I} \left\{\text{$\mathcal{G}_\gamma$ is a DAG}\right\}
    \end{align}
    where $p(\gamma_j|\mathcal{D}) = p(X_j|X_{\gamma_j}, \gamma_j) p^u(\gamma_j)$ and $p^u(\cdot)$ is the unconstrained prior
    \begin{align}
         p^u(\gamma_j) = \left(\frac{h}{1-h}\right)^{d_{\gamma_j}}.
    \end{align}
\end{property}

\section{The adaptive $\phi_t$ parameter}

\begin{figure}[t]
    \centering  \includegraphics[scale=0.4]{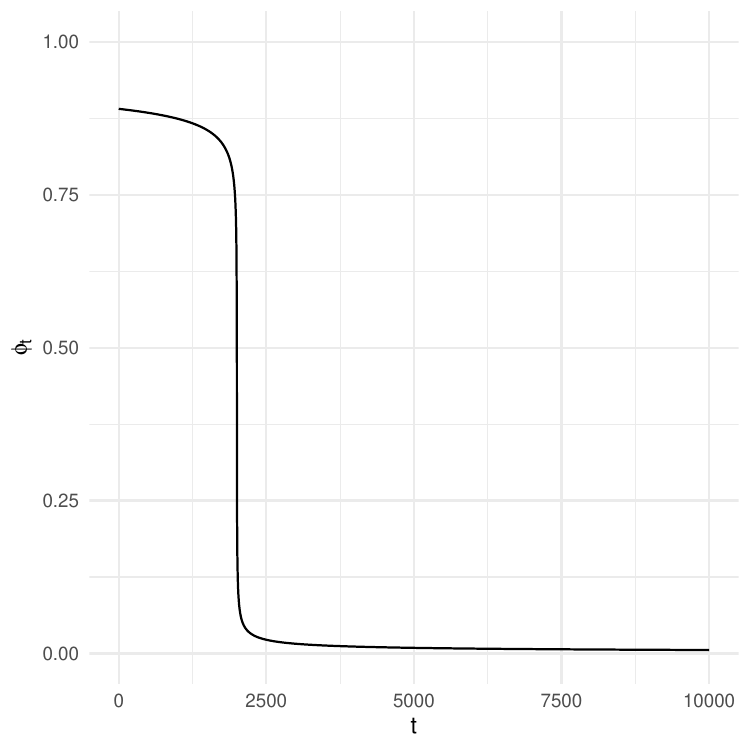}
    \caption{The plot of diminishing sequence $\phi_t$ for 10,000 iterations with burn-in period of 2,000 iterations.}
    \label{fig:phiplot}
\end{figure}

The decaying parameter $\phi_t$ is used for the adaptation of the algorithmic tuning parameter $\eta$ as given in (9), and controls the trade-off between the pre-processing approximation $\Tilde{\pi}$ and the ergodic average $\hat{\pi}^{(t)}$. Intuitively, one should rely more on the approximation $\Tilde{\pi}$ during the burn-in period, since the ergodic average $\hat{\pi}^{(t)}$ has not fully converged. After the burn-in period though, the proposal should rely more heavily on the ergodic average. Based on the arguments, we use the following sequence of $\phi_t$ to adapt $\eta$:
\begin{align}
    \phi_t = 
    \begin{cases}
        1 - \frac{1}{2} \left(\frac{1}{N_b - t + 1}\right)^{0.2} & \text{if $t \leq N_b$} \\
        \frac{1}{2} \left(\frac{1}{t - N_b}\right)^{0.5} & \text{if $t > N_b$.}
    \end{cases} \label{mat:phit}
\end{align}
In particular, we have $\phi_t = 1/2$ when $t = N_b$. An example plot of decaying $\phi_t$ is given in Figure \ref{fig:phiplot}, where we considered 10,000 iterations, with the first 2,000 iterations discarded as burn-in.

In addition, in the following lemma we derive the decaying rate of $\phi_t$, and thus of $\eta^{(t)}$ as well:

\begin{lemma}
    After the burn-in period, the diminishing rate of $\phi_t$ defined by (\ref{mat:phit}) is $\mathcal{O}(t^{-1})$. And thus the diminishing rate of tuning parameter $\hat{\eta}^{(t)}$ in (9) is $\mathcal{O}(t^{-1})$.
\end{lemma}

\begin{proof}
    Suppose $t > N_b$, we start from the definition of $\eta_t$ and consider the diminishing rate
    \begin{align*}
        |\phi_{t+1} - \phi_t| & = \bigg|\frac{1}{2}\left(\frac{1}{t+1-N_b}\right)^{0.5} - \frac{1}{2}\left(\frac{1}{t-N_b}\right)^{0.5}\bigg| \\
        & \leq \frac{1}{2} \left(\frac{(t+1-N_b)^{0.5} - (t-N_b)^{0.5} }{(t+1-N_b)^{0.5} (t-N_b)^{0.5}} \right) \\
        & \leq \frac{1}{2} \frac{1}{t-N_b} \\
        & = \mathcal{O}(t^{-1})
    \end{align*}
    as required.

    We can also show that the diminishing rate of the ergodic average $\hat{\pi}^{t}_{ij}$ has the same decaying rate
    \begin{align*}
        |\hat{\pi}_{ij}^{(t+1)} - \hat{\pi}_{ij}^{(t)}| & = \bigg| \frac{t}{t+1} \hat{\pi}_{ij}^{(t)} + \frac{1}{t+1} \gamma_{ij}^{(t+1)} - \hat{\pi}_{ij}^{(t)}  \bigg| \\
        & \leq \frac{1}{t+1} \hat{\eta}^{(t)}_{ij} + \frac{1}{t+1}\gamma_{ij}^{(t+1)} \\
        & \leq \frac{2}{t+1} = \mathcal{O}(t^{-1})
    \end{align*}
    as both $\gamma_{ij}^{(t)}$ and $\gamma_{ij}^{(t+1)}$ are bounded between 0 and 1.
    
    We complete the proof by showing that also the tuning parameter $\hat{\eta}_{ij}^{(t)}$ has the same decaying rate
    \begin{align*}
        |\hat{\eta}_{ij}^{(t+1)} - \hat{\eta}_{ij}^{(t)}|  & = |(\phi_t - \phi_{t+1})\Tilde{\pi}_{ij} + (\phi_{t+1} - \phi_{t})\hat{\pi}_{ij}^{(t+1)}| \\ 
        & \leq |\phi_t - \phi_{t+1}|\Tilde{\pi}_{ij} + |\phi_{t+1} - \phi_{t}| \hat{\pi}_{ij}^{(t+1)} \\
        & = \mathcal{O}(t^{-1}) + \mathcal{O}(t^{-1}) = \mathcal{O}(t^{-1})
    \end{align*}
    as $\Tilde{\pi}_{ij}$ is between 0 and 1.
    
\end{proof}

Given the above lemma, and combining Lemmas 1 and 2 in Section 4.2.3 of \cite{liang2022adaptive}, we can establish the similar ergodicity and strong law of large numbers as in Theorem 2 of \cite{liang2022adaptive}.

\section{The warm-start approximation $\Tilde{\pi}_{ij}$}
\label{apx:warmstart_approx}

The main idea of how to compute the warm-start approximation $\Tilde{\pi}_{ij}$ is explained in Section 3.2 of the main body. In this section, we will fill in with more details on calculation of $\Tilde{\pi}_{ij}$. 

Consider the extended collection of permissible parent set $\mathbf{h}^j_+$, formally defined as
\begin{align}
    \mathbf{h}^j_+ = \mathbf{h}^j \cup \left\{ m \cup \{j\} ~ | ~ m \in  \mathbf{h}^j ,  ~ j \in (h^j)^c \setminus\{j\} \right\}.
\end{align}
The extended set $\mathbf{h}^j_+$ includes the parent set implied by $\mathbf{h}^j$, but also the additional parent sets obtained by including at least one more parent not included in $h^j$. Trivially thus the extended set  $\mathbf{h}^j_+$ contains more permissible parents and minimises the risk of leaving out important parents when reducing the search space to a skeleton graph for scalability reasons \citep{kuipers_2022}.

We will show now how we calculate the marginal probability $\pi^u_{ij}$ under the unconstrained posterior $\pi^u$ and the extended sets $\{\mathbf{h}^j_+\}$. We define $\Gamma_j \subset \{0,1\}^p$ to be the sub-space of $\Gamma$ such that $\gamma_{jj} = 0$ if $\gamma_j \in \Gamma_j$. Given Markov property, thanks to which the columns of $\gamma$ can be factorised, we can define the unconstrained posterior distribution as
\begin{align}
    \pi^u(\gamma) \propto \prod_{j=1}^n p(\gamma_j|\mathcal{D}),
\end{align}
the marginal distribution of $\gamma_j$ under the unconstrained posterior $\pi^u$ is 
\begin{align}
    \pi^u(\gamma_j) = \frac{p(\gamma_j|\mathcal{D})}{\sum \limits_{\gamma_j \in \Gamma_j} p(\gamma_j|\mathcal{D})}.
\end{align}
However, the sub-space $\Gamma_j$ contains $2^{n-1}$ candidate parent sets and the full enumeration is computationally infeasible when $n$ is big. So we replace the subspace by the extended collection of parent sets $\mathbf{h}^j_+$ which contains much less models then $\Gamma_j$. So this approximate the above as
\begin{align}
    \pi^u(\gamma_j) \approx \frac{p(\gamma_j|\mathcal{D})}{\sum \limits_{m \in \mathbf{h}^j_+} p(\rho(m)|\mathcal{D})}.
\end{align}
where $\rho:[p] \to \{0,1\}^p$ is a one-to-one mapping that converts the parent set $m$ to $\gamma_j$ which lies in the binary space $\Gamma_j$. Then we approximate the PEP with the following formula:
\begin{align}
    \pi^u_{ij} & \stackrel{\text{def}}{=} \pi^u(\gamma_{ij} = 1) \approx \sum_{m \in \textbf{h}^j_+: ~ i \in m} \pi^u(\rho(m)) .
\end{align}
The set $\{m \in \textbf{h}^j_+~| ~ i \in m\}$ is equivalent to the set $\{\gamma_j \in \Gamma_j~| ~ \gamma_{ij} = 1\}$ of which $X_i$ must be a parent of $X_j$ from all these permissible parent sets. We also guarantee $\pi_{jj}^u = 0$ since $j$ is not one of the permissible parents of $j$ and not included in $\textbf{h}^j_+$.

After the approximations, the $\pi^u_{ij}$ are calculated and stored, then we can compute the warm-start approximations $\Tilde{\pi}_{ij}$, followed by the equations (11) and (12) described in the main body. The calculations of $\Tilde{\pi}_{ij}$ are equivalent to
\begin{align}
    \Tilde{\pi}_{ij} = \frac{\pi^u_{ij}(1-\pi^u_{ji})}{\pi^u_{ij}(1-\pi^u_{ji}) + (1-\pi^u_{ij})\pi^u_{ji} + (1-\pi^u_{ij})(1-\pi^u_{ji})}.
\end{align}
The last step combines $\Tilde{\pi}_{ij}$ with the ergodic average $\hat{\pi}_{ij}$ to form the algorithmic tuning parameters $\hat{\eta}_{ij}$ used in the PARNI-DAG proposal.

\section{The full PARNI-DAG proposal}
\label{apx:full_PARNI-DAG}

We first recall the random neighbourhood construction specified in Section 3.1 of the main body. The neighbourhood indicator $k \in \mathcal{K} = \{0,1\}^{n \times n}$ indicates the positions that can be potentially flipped in the current DAG $\gamma$, and $k$ follows the conditional distribution as given in (3) of the main body. The neighbourhood $\mathcal{N}(\gamma, k)$ given in (4) of the main body consists of $2^{d_k}$ DAG models, and the full enumeration over it is computationally efficient (and sometimes infeasible). For this reason we consider a pointwise implementation consisting in smaller sub-proposals. The first step in the pointwise implementation is to convert $k$ into a set of variables $K = \{K_r\}_{r=1}^R$, which can take two categories:
\begin{itemize}[leftmargin=5ex]
    \item if $k_{ij} = 1$ but $k_{ij} = 0$, $K_r$ only consists of one position $K_r = \{(i,j)\}$;
    \item if $k_{ij} = 1$ and $k_{ji} = 1$, $K_r$ then contains two positions $K_r = \{(i,j), ~(j,i)\}$.
\end{itemize}
The order of $K_r$ is random. To fully specify the sub-neighbourhoods $\mathcal{N}(\gamma(r-1), K_r)$ at each time $r$, we first define a new mapping that ``flips" the components of $\gamma$ according to the set $K$,
\begin{align}
    \gamma^\prime = \text{flip}(\gamma, K) ~, 
\end{align}
with $\gamma^\prime_{ij} = 1 - \gamma_{ij}$ if $(i,j) \in K$ and $\gamma^\prime_{ij} = \gamma_{ij}$ otherwise. The resulting neighbourhoods $N(\gamma(r-1), K_r)$ are defined as follows:
\begin{itemize}[leftmargin=5ex]
    \item If $K$ only contains one position $(i,j)$, then
    \begin{align}
        \mathcal{N}(\gamma(r-1), K_r) = \left\{ \gamma(r-1), ~ \text{flip}(\gamma(r-1), K_r) \right\}. \label{mat:neigh_1}
    \end{align}
    \item If $K$ contains two positions $(i,j)$ and $(j,i)$, we construct a neighbourhood with reversal move given by
    \begin{align}
        \mathcal{N}(\gamma(r-1), K_r) = \{ & \gamma(r-1), ~ \text{flip}(\gamma(r-1), (i,j)),   \notag \\
        & \text{flip}(\gamma(r-1), (j,i)), ~ \text{flip} (\gamma(r-1), K_r)\}. \label{mat:neigh_2}
    \end{align}
\end{itemize}

The next components in PARNI-DAG are the normalising constants of the sub-proposals $q_{g,K_r}$ and of the reversal sub-proposals $q_{g,K^\prime_r}$, used to facilitate the calculations of Metropolis-Hastings acceptance probability. As mentioned in the main body, we construct the auxiliary variables $K^\prime$ in the reversal move where $K^\prime$ consists of the same elements of $K$ but with different order. One property of such a design is that the intermediate DAGs in the reversal move are identical to the those ones in the proposal move but only with opposite order. We then can re-use the posterior model probabilities calculated during the proposal move to compute the reversal probabilities directly. In addition, the Proposition 3 of \cite{liang2022adaptive} implies that the acceptance probability in (7) can be expressed in terms of the product of the normalising constants. Given the normalising constants 
\begin{align}
    Z(r) & = \sum_{\gamma^* \in \mathcal{N}(\gamma(r-1), K_r)} g \left(\frac{\pi(\gamma^*)p(K_r|\gamma^*)}{\pi(\gamma(r-1))p(K_r|\gamma(r-1))}\right) \label{mat:normal_1}\\
    Z(R - r+1)^\prime & =  \sum_{\gamma^* \in \mathcal{N}(\gamma(r), K_r)} g \left(\frac{\pi(\gamma^*)p(K_r|\gamma^*)}{\pi(\gamma(r))p(K_r|\gamma(r))}\right),\label{mat:normal_2}
\end{align}
the MH acceptance probability is then simplified as follows
\begin{align}
    \alpha(\gamma,\gamma^\prime) =  \min \left\{1, \prod_{r=1}^R \frac{Z(r)}{Z^\prime(R - r + 1)} \right\}. \label{mat:PARNI-accprob}
\end{align}
To summarise the PARNI-DAG proposal, we now give its full algorithmic pseudo-code is given in \textbf{Algorithm} \ref{alg:PARNI}.

\begin{algorithm}[t]
\caption{The PARNI-DAG proposal} \label{alg:PARNI}
\setstretch{1.5}
\begin{algorithmic}
\State Compute $\{\Tilde{\pi}_{ij}\}$ as described in \textbf{Section 3.2} and \textbf{Appendix \ref{apx:warmstart_approx}}
\State Initialise $\gamma^{(1)}$, $\omega^{(1)}$ and $\hat{\eta}^{(1)} = \{\hat{\eta}_{ij}^{(1)}\}$
\For{$t = 1$ to $t = T$}
\State Sample $k \sim p_{\hat{\eta}^{(t)}}(~\cdot~|~\gamma^{(t)})$ as in (3)
\State Construct $K = \{K_r\}$ as described in \textbf{Appendix \ref{apx:full_PARNI-DAG}} 
\State Set $\gamma(0) = \gamma^{(i)}$ and $\Tilde{\mathcal{N}} = 0$
\For{$r = 1$ to $r = R$}
\If{$K_r$ has one element} 
\State Construct $\mathcal{N}(\gamma(r-1), K_r)$ as in (\ref{mat:neigh_1})
\Else
\State Construct $\mathcal{N}(\gamma(r-1), K_r)$ as in (\ref{mat:neigh_2})
\EndIf
\State Sample $U_1 \sim \text{Unif}(0,1)$
\If{$U_1 < \omega^{(t)}$}

\State Sample $\gamma(r) \sim q_{g,  K_r}(\gamma(r-1),~ \cdot~)$ as in (5)
\State Calculate $Z(r)$ as in (\ref{mat:normal_1})
\State Calculate $Z^\prime(R - r + 1)$ as in (\ref{mat:normal_2})
\Else
\State Set $\gamma(r) = \gamma(r-1)$, $Z(r) = Z(R - r + 1)^\prime = 1$.
\EndIf
\EndFor
\State Set $\gamma^\prime = \gamma(R)$, sample $U_2 \sim \text{Unif}(0,1)$ and compute $\alpha(\gamma^{(i)}, \gamma^\prime)$ as in (\ref{mat:PARNI-accprob});
\If{$U_2 < \alpha(\gamma^{(i)}, \gamma^\prime)$} 
\State Set $\gamma^{(t+1)} = \gamma^\prime$
\Else 
\State Set $\gamma^{(t+1)} = \gamma^{(t)}$
\EndIf
\State Update $\hat{\eta}^{(t+1)} = \{\hat{\eta}^{(t+1)}_{ij}\}$ according to (9).
\State Update $\omega^{(t+1)}$ according to (13).
\EndFor
\end{algorithmic}
\end{algorithm}

\section{Implementation of other MCMC algorithms}

In this section, we will briefly describe the Add-Delete-Reverse (ADR) proposal, the order MCMC and the partition MCMC implemented in the experimental Section 4.

\subsection{The Add-Delete-Reverse proposal}

The ADR proposal is another example of random neighbourhood proposal. After the random neighbourhood is generated, unlike the PARNI-DAG proposal, ADR uniformly proposes a model within that neighbourhood.

In ADR, there are three possible moves, ``Addition'', ``Deletion'' and ``Reversal'', which defines the space $\mathcal{K}$ of indicator $k$. The indicator $k$ is then uniformly drawn from one of these three possible moves. We construct the neighbourhood $\mathcal{N}(\gamma, k)$ according the the current DAG $\gamma$ and indicator $k$ as the following
\begin{align*} 
     \mathcal{N}(\gamma, \text{``Addition''})  & = \{\gamma^* \in \Gamma \, | \, d_H(\gamma, \gamma^*) = 1, \, d_{\gamma^*} = d_\gamma + 1, \text{$\gamma^*_{ji} = 0$ if $\gamma_{ij} = 1$} \} \\
     \mathcal{N}(\gamma, \text{``Deletion''}) &  = \{\gamma^* \in \Gamma \, | \, d_H(\gamma, \gamma^*) = 1, \, d_{\gamma^*} = d_\gamma - 1 \} \\
     \mathcal{N}(\gamma, \text{``Reversal''}) &  = \{\gamma^* \in \Gamma \, | \, d_H(\gamma, \gamma^*) = 2, \,  d_{\gamma^*} = d_\gamma, \, \text{$\exists$ $i,j$: $\gamma_{ij} = \gamma^*_{ji} = 0$ if $\gamma_{ji} = \gamma^*_{ij} = 1$} \}.
\end{align*}
where $d_H(\cdot,\cdot)$ denotes the Hamming distance and $d_\gamma$ is the number of edges induced by $\gamma$. A new DAG model $\gamma^\prime \in \Gamma$ is proposed according to the uniform density:
\begin{align}
    q^{\text{ADR}}_k (\gamma, \gamma^\prime) = 
    \begin{cases}
        \frac{1}{|\mathcal{N}(\gamma, k)|}, & \text{if $\gamma \in \mathcal{N}(\gamma, k)$} ,\\
        0, & \text{otherwise.}
    \end{cases}  \label{eq:ADRkern}
\end{align}
As ADR is not guaranteed (unlike PARNI-DAG) to return a DAG, we carry out acyclicity diagnostics check when $\gamma^\prime$ is sampled. If $\gamma$ does not lead to a valid DAG, we will accept the current state $\gamma$ as the next state of the Markov chain. If $\gamma^\prime$ leads to a valid DAG, to preserve $\pi$-reversibility, we consider $k^\prime$ in the reverse move where $k^\prime$ is the opposite move to $k$ (\textit{e.g.} the addition is opposite to the deletion move, and vice versa). To complete the ADR proposal, the new DAG $\gamma^\prime$ is accepted as the next state of the Markov chain with probability
\begin{align*}
    \alpha_k(\gamma,\gamma^\prime) & = \min\left\{1,\frac{\pi(\gamma^\prime)q_{k^\prime}(\gamma^\prime, \gamma)}{\pi(\gamma)q_{k}(\gamma, \gamma^\prime)}\right\} \\
    & = \min\left\{1,\frac{\pi(\gamma^\prime)|\mathcal{N}(\gamma,k)|}{\pi(\gamma)|\mathcal{N}(\gamma^\prime, k^\prime)|}\right\} . 
\end{align*}

\subsection{Order MCMC and Partition MCMC}

Order and Partition MCMC samplers rely on DAG scoring measures where $p(\mathcal{G} | \mathcal{D}) = \prod^n_{j=1} S( X_j, \text{Pa}(X_j) | \mathcal{D} ) $, and on restricting the search space of DAGs to some lower capacity set. \cite{friedman_2003} first proposed order MCMC, which operates in the space of orders $(\bm{X}, \prec)$ instead of the DAGs one. This search space is smoother, so that order MCMC generally achieves faster convergence with respect to classic neighborhood MCMC. The method considers permutations $(j_1, ..., j_n)$ of nodes that imply a linear order $j_1 \prec ... \prec j_n$. Each linear order is then given a score $R( \prec | \mathcal{D})$ equal to the sum of scores of all the compatible DAGs with the order $\prec \, \in (\bm{X}, \prec)$. To restrict the space capacity even more, sum and product operations can be swapped so that the order's score is:
\begin{equation*}  \small
\begin{split}
    R( \prec | \mathcal{D}) = \sum_{\mathcal{G} \in \prec} p(\mathcal{G} | \mathcal{D}) & \propto \sum_{\mathcal{G} \in \prec} \prod^n_{j=1} S( X_j, \text{Pa}(X_j) | \mathcal{D} )  \\
    & \propto \prod^n_{j=1} \sum_{\text{Pa} (X_j) \in \prec } S( X_j, \text{Pa}(X_j) | \mathcal{D} ) \, ,
\end{split}
\end{equation*}
where $S(\cdot)$ is a DAG scoring function, such as the Bayesian Gaussian equivalent (BGe) score for continuous variables \citep{geiger2002parameter}. Constructing a MCMC in the space of orders means moving from order $\prec$ to order $\prec'$ with MH acceptance probability $\alpha_{\prec} ( \prec, \prec' ) = \min \left\{ 1, \frac{R(\prec^\prime | \mathcal{D})}{R(\prec | \mathcal{D})} \right\}$, according to following moves: a) ``Local Swap" swaps adjacent nodes in $\prec$; b) ``Global Swap" swaps random nodes in $\prec$; c) ``Relocate" picks a node and replaces it in all possible positions in $\prec$, while keeping the rest of the order unchanged. Order MCMC is efficient, but has two main drawbacks compared to neighborhood-like structure MCMC: i) it outputs a sample of orders, not a DAG directly; sample of DAGs can be obtained by sampling, for each node independently, a parent set compatible with the relative order $\prec$, thanks to score decomposability (then get the DAG with maximum score in the sample as MAP estimate); ii) Order MCMC places a non-uniform prior on some DAGs by over-representing them in more than one order $\prec$, which causes the sample to be biased \citep{ellis2008learning}. 

In the attempt to correct sampling bias, \cite{kuipers_2017} proposes Partition MCMC, which operates in the space of ordered partitions instead. A labelled partition $\Lambda$ is a pair of order $\prec$ and partitioning vector $\bm{\delta} = (\delta_1, ..., \delta_p)$ where $p < n$, and $\sum^p_{j=1} \delta_j = n$. Vector $\bm{\delta}$ divides the order into $p$ parts $v_1, ..., v_p$ such that $v_1$ contains the first $\delta_1$ nodes of the order $\prec$ and so on. A DAG is compatible with partition $\Lambda = (\prec, \bm{\delta})$ if for every node $X_j \in v_j$: i) $X_j$ has at least one parent in part $v_{j+1}$; ii) all $\text{Pa}(X_j)$ belong to parts with indices higher than $j$; iii) $\text{Pa}(X_j) = \emptyset$ only if $j=p$. The possible moves of partition MCMC are: a) swap 2 nodes from different parts; b) swap 2 nodes from adjacent parts; c) split or join two parts; d) move a node in another existing part or create a new part with that node. Partition MCMC is unbiased in terms of DAG sampling; however it is extremely slow as it has very high computational complexity, and therefore can only be used to sample DAGs with very few nodes \citep{kuipers_2022}.

The orderMCMC and partitionMCMC schemes in the experiments are implemented via the \texttt{BiDAG}\footnote{More information on: \url{https://cran.r-project.org/web/packages/BiDAG/BiDAG.pdf}.} package \citep{suter2021bayesian} (available from \textbf{CRAN}) written in \texttt{rcpp} \citep{Eddelbuettel2011rcpp}.

\section{Additional experimental results}

\subsection{Addition to Section 5.1}

\begin{figure}[t]
     \centering
     \begin{subfigure}[b]{0.3\textwidth}
         \centering
         \includegraphics[width=\textwidth]{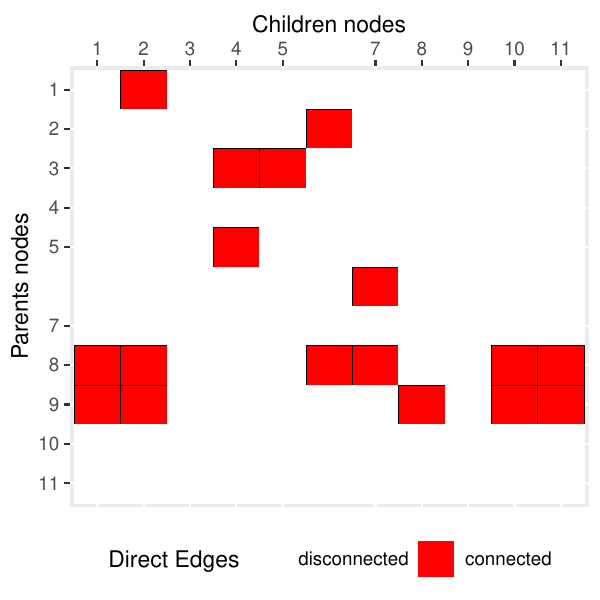}
         \caption{}
         \label{fig:protein_gt}
     \end{subfigure}
     \hfill
     \begin{subfigure}[b]{0.3\textwidth}
         \centering
         \includegraphics[width=\textwidth]{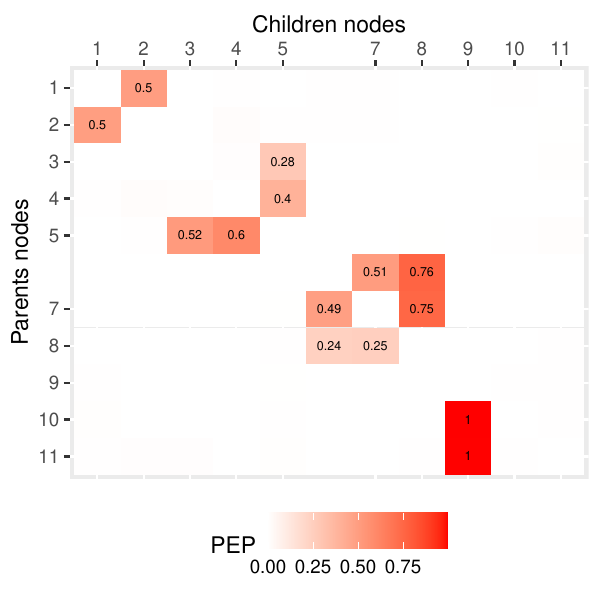}
         \caption{}
         \label{fig:protein_truePEP}
     \end{subfigure}
     \hfill
     \begin{subfigure}[b]{0.3\textwidth}
         \centering
         \includegraphics[width=\textwidth]{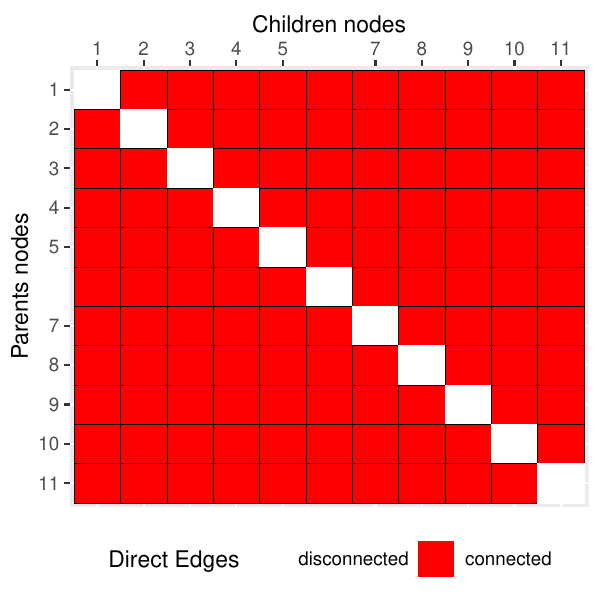}
         \caption{}
         \label{fig:protein_fullskel}
     \end{subfigure}
        \caption{Protein dataset: (a) The ground-truth DAG. (b) The true PEPs estimated by a long chain of partition MCMC algorithm. (c) The full skeleton used in Section 5.1.}
        \label{fig:protein_graphs}
\end{figure}

\paragraph{Protein data.} We provide additional information on the real-world protein-signalling dataset ($n = 11$, $N = 853$) studied in \textbf{Section 4.1}. The ground-truth DAG constructed through expert knowledge is provided in Figure \ref{fig:protein_graphs} (a). The ground-truth estimate of PEP estimated by running a long chain of the partition MCMC algorithm is provided in Figure \ref{fig:protein_graphs} (b). The full skeleton used to implement different MCMC schemes is provided in \ref{fig:protein_graphs} (c).

\begin{figure}[t]
     \centering
     \begin{subfigure}[b]{0.475\textwidth}
         \centering
         \includegraphics[width=\textwidth]{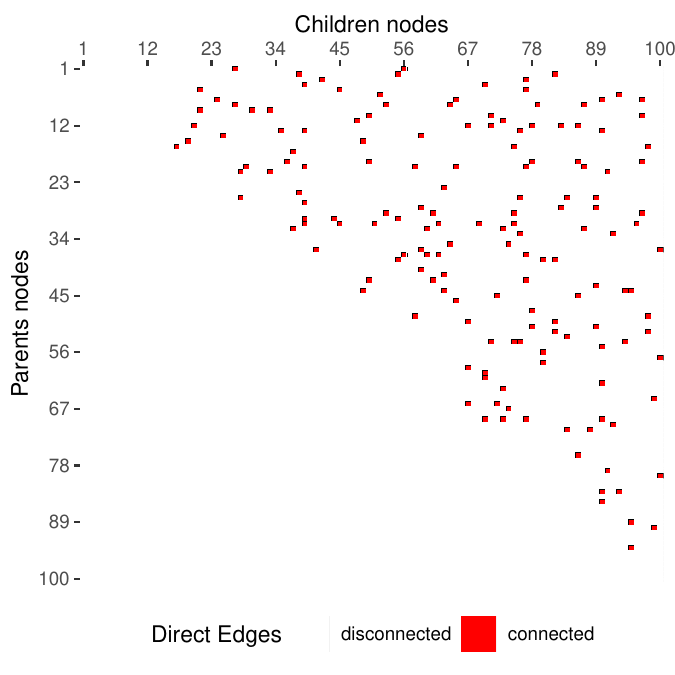}
         \caption{}
         \label{fig:gsim100_gt}
     \end{subfigure}
     \hfill
     \begin{subfigure}[b]{0.475\textwidth}
         \centering
         \includegraphics[width=\textwidth]{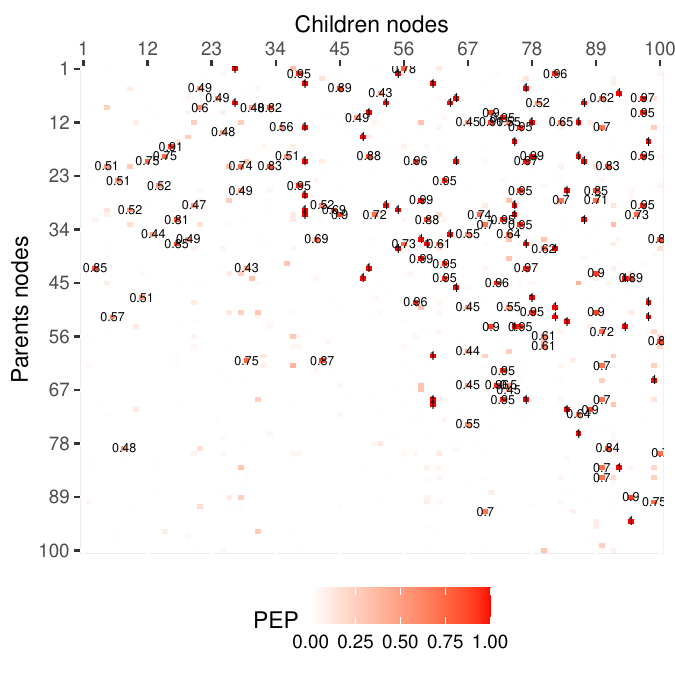}
         \caption{}
         \label{fig:gsim100_truePEP}
     \end{subfigure}
     \hfill
     \vskip\baselineskip
    \begin{subfigure}[b]{0.475\textwidth}
         \centering
         \includegraphics[width=\textwidth]{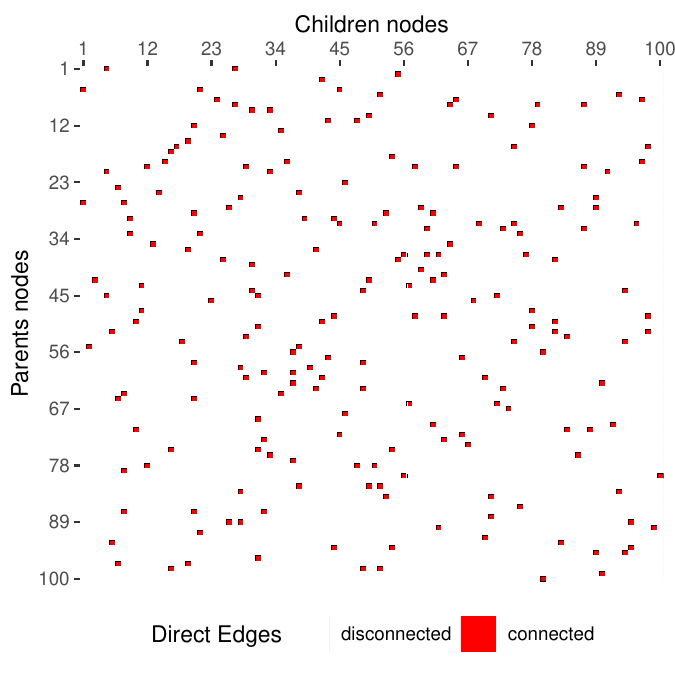}
         \caption{$\mathcal{H}_{\text{PC}}$}
         \label{fig:gsim100_skelPC005}
     \end{subfigure}
     \hfill
     \begin{subfigure}[b]{0.475\textwidth}
         \centering
         \includegraphics[width=\textwidth]{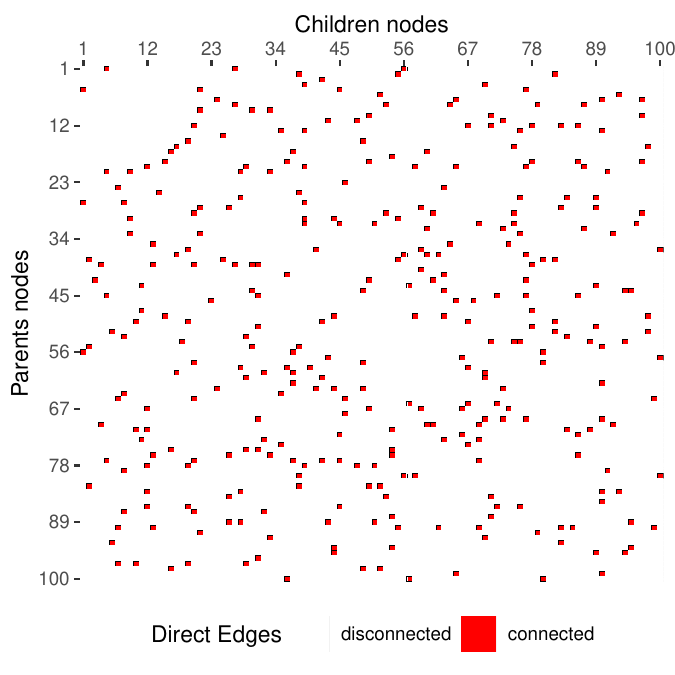}
         \caption{$\mathcal{H}_{\text{iter}}$}
         \label{fig:gsim100_skeliter}
     \end{subfigure}
    \caption{\texttt{gsim100} dataset: (a) The ground-truth DAG. (b) The true PEPs estimated by a long chain of the PARNI-DAG proposal. (c) $\mathcal{H}_{\text{PC}}$ with significance level of 0.05. (d) $\mathcal{H}_{\text{iter}}$. }
        \label{fig:gsim100_graphs}
\end{figure}

\paragraph{gsim100 data.}

We also provide additional information on the simulated \texttt{gsim100} dataset ($n = 100$, $N = 100$) studied in \textbf{Section 4.1}, and featured in \cite{suter2021bayesian}. The ground-truth DAG, containing 161 true edges, used to simulate data is presented in Figure \ref{fig:gsim100_graphs} (a). The ground-truth PEP estimates obtained from a long chain run of the PARNI-DAG proposal are provided in \ref{fig:gsim100_graphs} (b). The skeleton learned via the PC-algorithm (with a significance rate of $5\%$) $\mathcal{H}_{\text{PC}}$ and the skeleton learned from the iterative MCMC $\mathcal{H}_{\text{iter}}$ are provided in Figure \ref{fig:gsim100_graphs} (c) and (d) respectively.

\subsection{Additional studies on the effects of tuning parameter of the PC-algorithm}

\begin{figure}[t]
     \centering
     \begin{subfigure}[b]{0.3\textwidth}
         \centering
         \includegraphics[width=\textwidth]{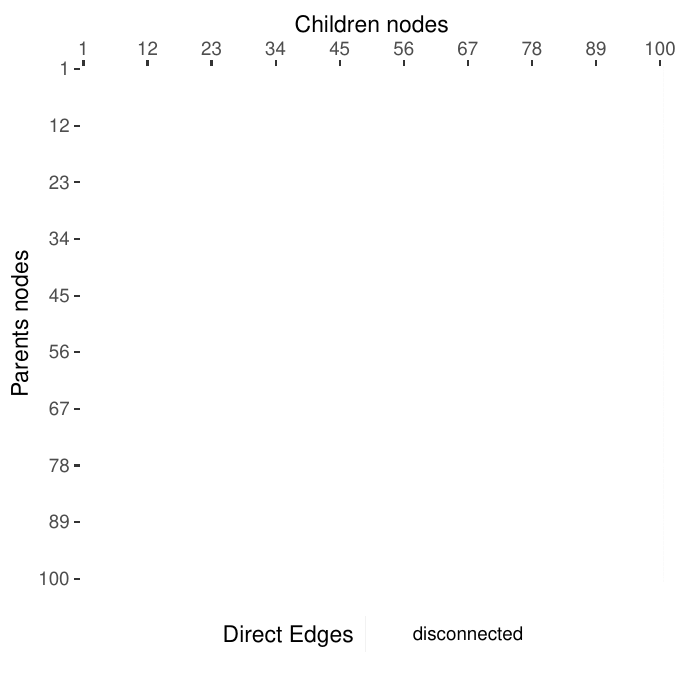}
         \caption{}
         \label{fig:gsim_PC00}
     \end{subfigure}
     \hfill
     \begin{subfigure}[b]{0.3\textwidth}
         \centering
         \includegraphics[width=\textwidth]{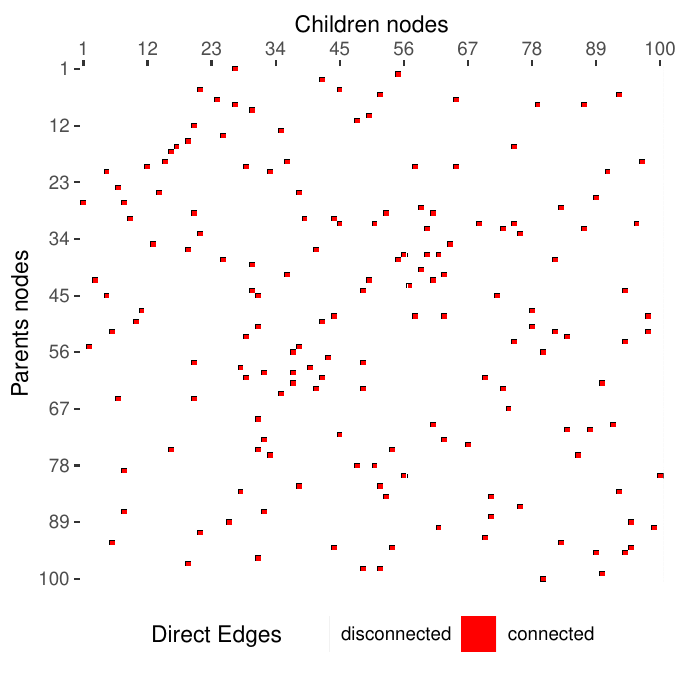}
         \caption{}
         \label{fig:gsim_PC001}
     \end{subfigure}
     \hfill
     \begin{subfigure}[b]{0.3\textwidth}
         \centering
         \includegraphics[width=\textwidth]{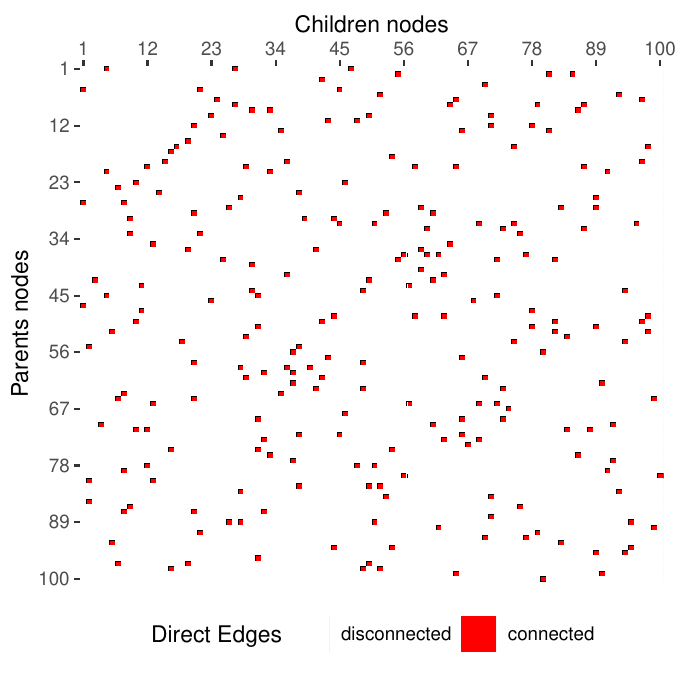}
         \caption{}
         \label{fig:gsim_PC01}
     \end{subfigure}
        \caption{\texttt{gsim100} dataset: (a) $\mathcal{H}_\text{PC}$ with significance level of 0. (b) $\mathcal{H}_\text{PC}$ with significance level of 0.01. (c) $\mathcal{H}_\text{PC}$ with significance level of 0.1}
        \label{fig:gsim100_skeleton_graphs}
\end{figure}

\begin{figure}[t]
    \centering
    \includegraphics[width = \textwidth]{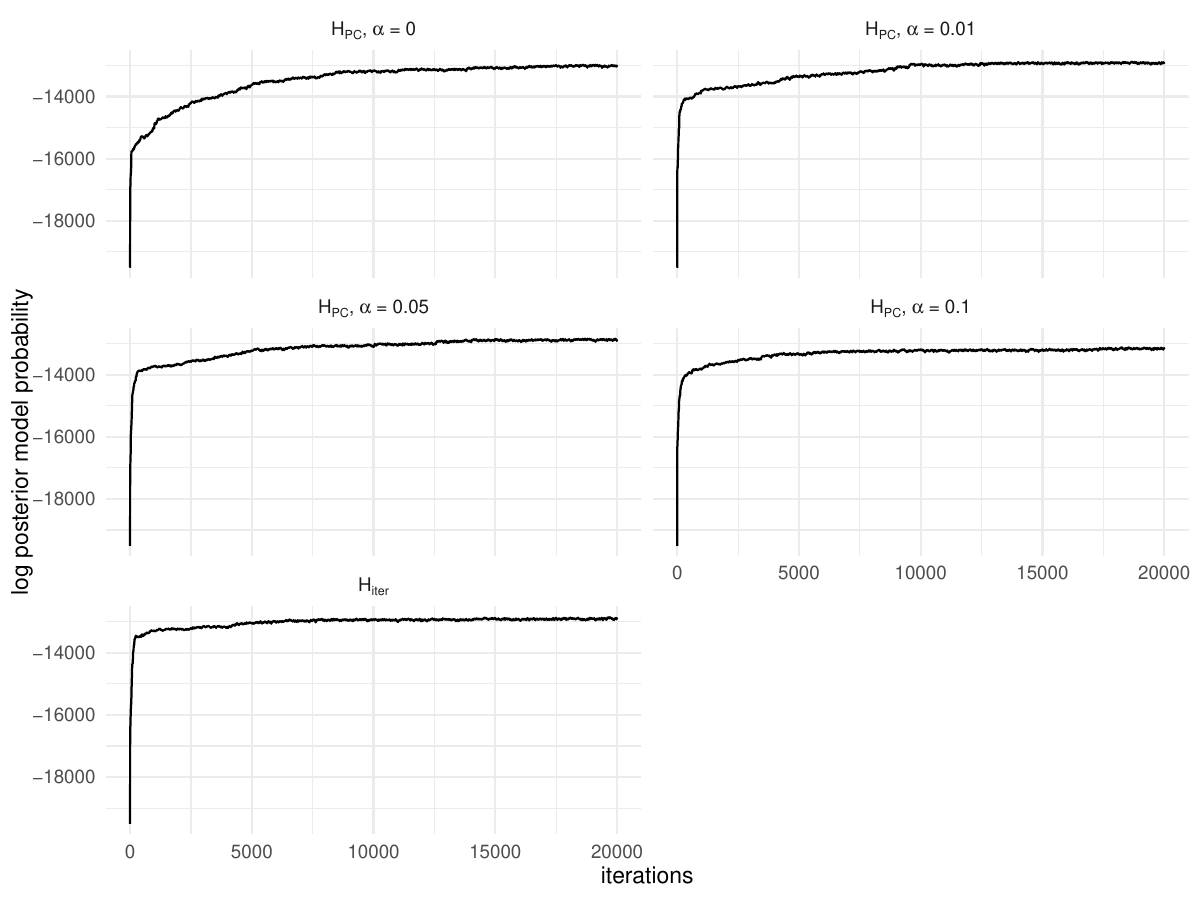}
    \caption{\texttt{gsim100} dataset: trace plots of log posterior model probabilities for the PARNI-DAG proposal under different skeletons.} \label{fig:gsim100_skeltrace}
\end{figure}

\begin{table}[t]
    \centering
    \ra{1.2}
    \begin{tabular}{c|c|c}
    \toprule
    skeleton  & significance rate ($\alpha$)   & MSE \\ 
    \midrule
    \multirow{4}{*}{$\mathcal{H}_{\text{PC}}$}
     & 0 & 2.40
    \\
     & 0.01 & 1.92 \\
     & 0.05 & 1.95 \\
     & 0.1 & 1.95 \\
    \midrule
    $\mathcal{H}_{\text{iter}}$ & - &  1.85 \\
    \bottomrule 
    \end{tabular}
    \caption{\texttt{gsim100} dataset: time-normalised median MSE ($\times 10^{-7}$) on estimating posterior edge probabilities. (\textit{Lower is better.})}
    \label{tab2:gsim_100}
\end{table}

In addition to \textbf{Section 4.1} results, we perform another batch of numerical studies to study the PARNI-DAG proposal run over different starting skeletons, by comparing their convergence behavior. The skeletons are mainly obtained by running the PC-algorithm with different levels of significance. We consider 4 different values of significance level, 0, 0.01, 0.05 and 0.1. The resulting skeletons are provided in Figure \ref{fig:gsim100_skeleton_graphs} (The skeleton with significance level of 0.05 is provided in Figure \ref{fig:gsim100_graphs} (c)). As the significance level of the PC-algorithm increases, the resulting skeleton includes more edges whilst the skeleton is empty when significance level is 0. In addition, we also compare these results to the skeleton from iteration MCMC as in \textbf{Section 4.1}.

Figure \ref{fig:gsim100_skeltrace} presents the trace plots of the log posterior model probabilities from the PARNI-DAG proposal under different skeletons. Using the skeleton $\mathcal{H}_{\text{PC}}$ with significance level of 0 trivially leads to the slowest convergence rate, while the skeleton $\mathcal{H}_{\text{iter}}$ (which includes more possible parent sets) results in the fastest convergence. The behaviour of the chains is similar. We can draw similar conclusions from the median MSEs presented in Table \ref{tab2:gsim_100}. The skeleton $\mathcal{H}_{\text{iter}}$ leads to the minimum MSE, while $\mathcal{H}_{\text{PC}}$ with significance level of 0 again trivially results in the largest MSE. The MSEs from other three options of significance levels are not significant different between each other, so using any one of them will give similar level of accuracy.

\subsection{Additional studies on DAG learning performance}

In addition to the results provided in Section 4.2 of the main body, in a similar way we compare PARNI-DAG with ADR and the same other methods in terms of accuracy in recovering the underlying DAG structure on an another example. We consider a randomly generated DAG structure featuring 50 nodes and 119 edges. Given this DAG structure, we simulate Gaussian data for three different sample sizes $N \in \{50, 100, 150\}$.

To recall, the models compared are: i) PC algorithm \citep{spirtes_2000_book}; ii) GES algorithm \citep{chickering_1996, chickering_2002}; iii) Order MCMC \citep{friedman_2003}; iv) Iterative MCMC \citep{kuipers_2022}; v) Partition MCMC \citep{kuipers_2017}; vi) ADR; vii) PARNI-DAG. 

\begin{figure}[t]
    \hspace{-0.9cm}
    \includegraphics[scale=0.44]{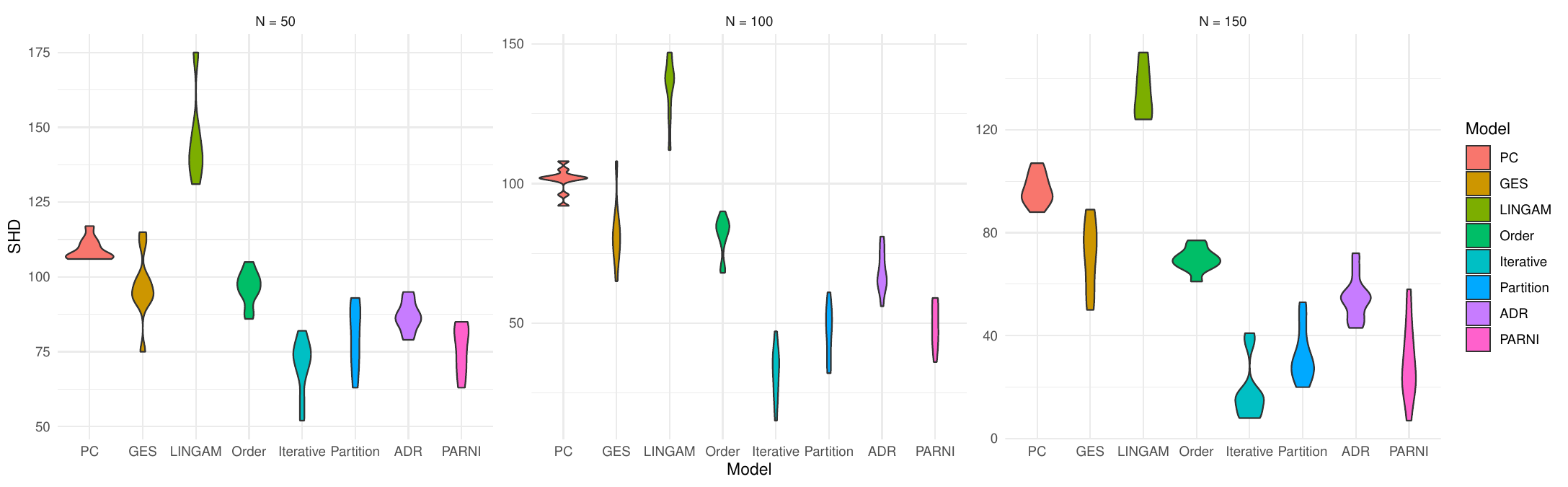}
    \caption{Distribution of SHD of each compared model, for different sample sizes $N \in \{50, 100, 150\}$, over $10$ replications.}
    \label{fig:violin}
\end{figure}

Models' performance is compared according to same metric, the \emph{Structural Hamming Distance} $d_H (\hat{\gamma}, \gamma)$ between the estimated and the true DAG. The experiments are replicated for $10$ times, for each of the sample sizes above. Results on the distributions of SHD are depicted in Figure \ref{fig:violin}. We can see how Iterative MCMC and PARNI-DAG are the best competing models out of this example. Note that this is a medium size graph, and while we have argued in the main body that PARNI-DAG particularly excels in settings characterized by larger size graphs ($|\mathcal{V}| \geq N$), it is nonetheless competitive with the best performing method, Iterative MCMC here, also on lower dimensional DAGs.

\clearpage

\bibliography{Refs}
